\documentclass[preprint,3p,times, 11pt, a4paper]{elsarticle}

\usepackage{url}
\usepackage[T1]{fontenc}

\usepackage[hidelinks]{hyperref}  
\usepackage{amsmath,amssymb,amsfonts,mathrsfs,mathtools,mathrsfs}
\usepackage{cleveref}
\usepackage{natbib}

\usepackage{subcaption}
\usepackage{graphicx}
\graphicspath{{Figures/}{Figures/BoxPlots}{Figures/ISA}}

\usepackage{listings}
\usepackage{array}
\usepackage{booktabs}

\usepackage{todonotes}

\usepackage{bm}

\newcommand{\ra}[1]{\renewcommand{\arraystretch}{#1}}

\usepackage{enumitem}
\newlist{paraenum}{enumerate*}{1}
\setlist[paraenum]{label=(\emph{\roman*})}

\usepackage{tikz}
\usetikzlibrary{shapes, arrows, positioning, matrix, fit}

\definecolor{purple}{rgb}{0.6314,0.5412,0.8941}
\tikzstyle{block}=[rectangle, minimum width=2cm, text width=2cm, minimum height=0.5cm, text centered, draw, thick]
\tikzstyle{arrow}=[thick, ->, >=stealth, draw]

\newcommand{\new}[1]{#1}

\usepackage{multirow}

\usepackage{setspace}

\makeatletter
\def\ps@pprintTitle{%
    \let\@oddhead\@empty
    \let\@evenhead\@empty
    \def\@oddfoot{\footnotesize%
        \begin{minipage}[t]{\textwidth}
            \raggedright
           \textit{This is the peer-reviewed author-version of \url{https://doi.org/10.1016/j.ejor.2024.06.005}, published in the European Journal of Operational Research. \textcopyright\ 2024. This manuscript version is made available under the LCC-BY-NC-ND 4.0 license \url{https://creativecommons.org/licenses/by-nc-nd/4.0/}.}
        \end{minipage}%
    }%
    \let\@evenfoot\@oddfoot
}
\makeatother

\usepackage{fancyhdr}
\fancypagestyle{mycustomstyle}{
    \fancyhf{}  
     \fancyhead[R]{%
        \begin{tikzpicture}[overlay, remember picture]
            \node[anchor=north west, inner sep=0pt, xshift=60pt, yshift=-50pt, text width=400pt] at (current page.north west) {\textsl{This is the peer-reviewed author-version of \url{https://doi.org/10.1016/j.ejor.2024.06.005}, published in the European Journal of Operational Research. \textcopyright\ 2024.}};
            \node[anchor=north east, inner sep=0pt, xshift=-65pt, yshift=-50pt] at (current page.north east) {\href{https://creativecommons.org/licenses/by-nc-nd/4.0/}{\includegraphics[width=2cm]{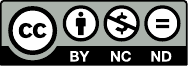}}}; 
        \end{tikzpicture}%
    }
    \fancyfoot[C]{\thepage} 
}

\begin{document}
\pagestyle{mycustomstyle}
\begin{frontmatter}

\author[Ghent,CVAMO]{David Van Bulck\corref{cor1}}
\ead{david.vanbulck@ugent.be}
\author[Ghent,CVAMO]{Dries Goossens}
\ead{dries.goossens@ugent.be}
\author[Zuse]{Jan-Patrick Clarner}
\author[Angelos]{Angelos Dimitsas}
\ead{a.dimitsas@uoi.gr}
\author[George]{George H.\ G.\ Fonseca}
\ead{george@ufop.edu.br}
\author[Carlos]{Carlos Lamas-Fernandez}
\ead{C.Lamas-Fernandez@soton.ac.uk}
\author[Martin]{Martin Mariusz Lester}
\ead{m.lester@reading.ac.uk}
\author[Zuse]{Jaap Pedersen}
\ead{pedersen@zib.de}
\author[Antony]{Antony E. Phillips}
\ead{aphi.xc@gmail.com}
\author[Roberto]{Roberto Maria Rosati}
\ead{robertomaria.rosati@uniud.it}

\cortext[cor1]{Corresponding author}

\affiliation[Ghent]{organization={Faculty of Economics and Business Administration, Ghent University},
            addressline={Tweekerkenstraat 2}, 
            city={Ghent},
            postcode={9000}, 
            country={Belgium}
    }
\affiliation[CVAMO]{organization={Core lab CVAMO, FlandersMake@UGent},
            country={Belgium}
    }
\affiliation[Zuse]{organization={Applied Algorithmic Intelligence Methods Department, Zuse Institute Berlin},
            addressline={Takustraße 7}, 
            city={Berlin},
            postcode={14195}, 
            country={Germany}
    }
\affiliation[Angelos]{organization={Department of Informatics, University of Ioannina},
            addressline={University of Ioannina}, 
            country={Greece}
    }
    \affiliation[George]{organization={Computing and Systems Department, Federal University of Ouro Preto},
            addressline={R. Diogo de Vasconcelos, 122}, 
            city={Ouro Preto},
            country={Brazil}
    }
    \affiliation[Carlos]{organization={CORMSIS (Centre for Operational Research, Management Science \& Information Systems), Southampton Business School, University of Southampton},
            country={UK}
    }
    \affiliation[Martin]{organization={Department of Computer Science, University of Reading},
            city={Reading},
            country={UK}
    }
    \affiliation[Antony]{organization={7bridges, www.the7bridges.com},
            addressline={23 Meard St}, 
            city={London},
            country={UK}
    }
    \affiliation[Roberto]{organization={DPIA, University of Udine},
            addressline={via delle Scienze 206}, 
            city={Udine},
            postcode={33100}, 
            country={Italy}
    }
\begin{abstract}
Any sports competition needs a timetable, specifying when and where teams meet each other. The recent International Timetabling Competition (ITC2021) on sports timetabling showed that, although it is possible to develop general algorithms, the performance of each algorithm varies considerably over the problem instances. 
This paper provides a problem type analysis for sports timetabling, resulting in powerful insights into the strengths and weaknesses of eight state-of-the-art algorithms. 
Based on machine learning techniques, we propose an algorithm selection system that predicts which algorithm is likely to perform best based on the type of competition and constraints being used (i.e., the problem type) in a given sports timetabling problem instance.
Furthermore, we visualize how the problem type relates to algorithm performance, providing insights and possibilities to further enhance several algorithms.
Finally, we assess the empirical hardness of the instances.
Our results are based on large computational experiments involving about 50 years of CPU time on more than 500 newly generated problem instances.
\end{abstract}

\title{Which algorithm to select in sports timetabling?}

\begin{keyword}


        OR in Sports\sep Sports scheduling \sep ITC2021 \sep Algorithm selection \sep Instance space analysis
\end{keyword}
\end{frontmatter}
\pagestyle{mycustomstyle}

\section{Introduction}

In sports timetabling, we need to assign a given set of matches to rounds (also called time slots) such that each team plays at most once per round. This problem is complicated by a diverse set of (sometimes conflicting) wishes and requirements that need to be taken into account. While sports timetabling problems are typically computationally highly demanding, this is not overly problematic in practice as tournament organizers face this problem only a few times per year, and generally have ample time to come up with a solution.

For a long time, most real-life sports tournaments have been scheduled by hand. During the last two decades, sports timetabling research yielded several dedicated algorithms, and the publication of numerous successful real-life applications (see e.g., \citet{Goossens2009,Ribeiro2012b,Recalde2013,Westphal2014,VanBulck2018, Duran2020a}). The performance of these algorithms was typically evaluated by comparing one or two timetables to those manually created by expert practitioners. There are two issues with this practice (see also \citet{Ceschia2022}). First, it only assesses the algorithm's performance under the specifics of one particular tournament, so that it is difficult to gain general insights on its performance with regard to other tournaments. Secondly, as the performance is not directly compared against other sports timetabling algorithms, it is hard to tell how well an algorithm truly performs.

Until recently, one of the main difficulties for performance assessment of sports timetabling algorithms was the lack of a uniform way to describe constraints and objectives in a unified instance and solution file format (with the travelling tournament problem being a notable exception, see \citet{Easton2001}). With the introduction of RobinX, which consists of a three-field classification and XML file templates to describe sports timetabling problem instances, \citet{VanBulck2019} overcame this issue. \citet{VanBulck2022b} used the RobinX format in the 2021 edition of the International Timetabling Competition (ITC2021), which focused on sports timetabling. 
This competition resulted in the development of several general sports timetabling solvers, able to handle the wide variety of constraints that one encounters in practice. Although general solution strategies (e.g., first-break-then-schedule, see \citet{Nemhauser1998}) had been proposed in the literature before, the development of general sports timetabling solvers, tested on a common benchmark of problem instances, advanced the field considerably.

The most important insight from ITC2021 is perhaps that it is possible to move away from tournament-specific algorithms to more generally-applicable solvers.
On the other hand, using a single algorithm for all types of applications is not the best choice either, as ITC2021 shows that different algorithms tend to work best on different types of problem instances.
This raises the question of which algorithm from the literature a practitioner should adopt to tackle the timetabling problem for their particular tournament.
This paper aims to answer that question by employing algorithm selection techniques predicting the best performing algorithm based on measurable properties (i.e., features) of problem instances.
Distinct from most algorithm selection research that requires a detailed specification of the problem instance, the features considered in this paper are based on the problem type, in the sense that they only involve specifying the type of competition and constraints being used.
Problem-type features are in line with the RobinX framework as well as the way in which the ITC2021 instances were generated, and are meaningful from a practitioner's point of view as they typically need to decide on purchasing a (single) solver (or service) before knowing the full details of the problem instances to be solved in future years.
Visualizing how the performance predictions relate to the problem type, a technique we coin problem type analysis and which is closely related to instance space analysis (see \citet{Smith-Miles2023}), we reveal the strength and weakness of eight state-of-the-art algorithms from the literature on ITC2021. Our insights are based on large computational experiments, involving over 500 newly generated sports timetabling problem instances.

The remainder of this paper reads as follows. 
\Cref{sec:problemForm} provides a description of the type of sports timetabling problems envisioned in this paper.
Next, \Cref{sec:framework} provides an overview on the literature on algorithm selection and explains how we adapt the framework of instance space analysis to work on the level of problem types.
\Cref{sec:metadata} provides an overview of the features, problem instances, and algorithms we collected for this research, and \Cref{sec:problemTypeAnalysis} visualizes and analyses the algorithm performances.
Conclusions and opportunities for further research are presented in \Cref{sec:conclusion}.

\section{The ITC2021 sports timetabling problem}\label{sec:problemForm}

We envision the sports timetabling problem from the ITC2021 competition, which requires constructing a double round-robin tournament (2RR), meaning that each team plays against every other team exactly once at home and once away.
Moreover, it assumes that the number of teams $n$ is even and that the timetable is compact (i.e., there are $2n-2$ time slots, or in other words, each team plays exactly one game per time slot).
Furthermore, the ITC2021 problem instances feature nine (somewhat simplified) constraint types from the classification framework by \citet{VanBulck2019}.
Constraints are either hard or soft, where hard constraints represent fundamental properties of the timetable that can never be violated and soft constraints represent preferences that should be satisfied whenever possible.
The objective in the ITC2021 problem instances is to minimize the overall (weighted) sum of deviations from violated soft constraints while respecting all hard constraints.
In the remainder of this section, we describe each of the constraint types (grouped into four constraint classes), and we refer to \citet{VanBulck2022b} for a more detailed description of the constraints and their representation in the competition file format of RobinX.

\begin{description}
	\item[\bf Capacity constraints] The first class of constraints regulates when teams can play home or away.\\[3pt]
		$\text{CA1}$ constraints limit the total number of home or away games played by a given team during a given set of time slots, regardless of the opponent.\\[3pt]
		$\text{CA2}$ constraints limit the total number of home or away games played by a given team against a given set of opponents during a given set of time slots.\\[3pt]
		$\text{CA3}$ constraints limit the total number of home or away games played by a given team against a given set of other teams during each sequence of a given number of time slots.\\[3pt]
		$\text{CA4}$ constraints limit the overall number of home games (away games, or games) played by a given set of teams against a given set of other teams during a given set of time slots.
	\item[\bf Break constraints] If a team plays two consecutive home games, or two consecutive away games, we say it has a break.\\[3pt]
		$\text{BR1}$ constraints limit the total number of breaks for a given team during a given set of time slots.\\[3pt]
		$\text{BR2}$ constraints limit the total number of breaks in the timetable. At most one constraint of this type (hard or soft) is present.
	\item[\bf Game constraints] The third class consists of only one constraint type.\\[3pt]
		$\text{GA1}$ constraints enforce or forbid the assignment of a (set of) game(s) to one or more specific time slots.
	\item[\bf Fairness and separation constraints] The last constraints aim to increase the fairness and attractiveness of the tournament, and are always soft. Moreover, there is at most one constraint of each type.\\[3pt]
		The $\text{FA2}$ constraint expresses the preference that the timetable is 2-ranking-balanced, meaning that the difference in the number of played home games between any two teams at any point in time is at most 2.\\[3pt]
		The $\text{SE1}$ constraint specifies that there are at least 10 time slots between any two games involving the same opponents.
\end{description}

Apart from these constraints, some problem instances also require that the timetable is `phased', meaning that each team meets every other team exactly once in the first $n-1$ time slots and once in the last $n-1$ time slots.

\section{Methodological framework}
\label{sec:framework}

As early as 1975, the renowned computer scientist Donald Knuth observed that algorithm performance on seemingly similar problem instances may strongly vary.
Nevertheless, predicting algorithm performance is far from trivial: `One of the chief difficulties associated with [...] combinatorial problems has been our inability to predict the efficiency of a given algorithm, or to compare the efficiencies of different approaches, without actually writing and running the programs' (see \citet{Knuth1975}).
Merely a year later, \citet{Rice1976} published a seminal work on algorithm selection, formally defining the problem as learning a mapping of measurable properties or characteristics of problem instances (i.e., features) to algorithms so as to select the best possible algorithm for each problem instance.
At the time of Rice's formulation of the algorithm selection problem, machine learning techniques were still in their early stages of development.
With the rise of machine learning techniques around 2000, the algorithm selection problem was `rediscovered' and suddenly received considerable attention in the literature (for an overview of this literature, see \citet{Kotthoff2016,Kerschke2019}).

Rice's Algorithm Selection framework can be used to predict \emph{when} algorithms perform well or \emph{poorly}, but does not delve into the question \emph{why} this is the case (see also \citet{Kotthoff2016}).
Moreover, even though Rice recognizes the importance of the quality of the available problem instances and features for the selection mapping to generalize well towards unseen data, his framework does not provide a systematic approach of assessing the quality of the metadata.
In order to overcome the aforementioned challenges, Smith-Miles and co-authors embed the algorithm selection problem in what they call Instance Space Analysis (ISA, see \citet{Smith-Miles2023}).
The distinguishing property of ISA is the use of a customised dimensionality reduction technique to project problem instances represented by their feature vector in a so-called high-dimensional feature space into a two-dimensional (2D) instance space.
One of the key insights is that a set of problem instances is representative for the broader population of the problem class if the problem instances are well scattered in the 2D space.
Moreover, by superimposing the performance of algorithms in this 2D space, the regions where an algorithm excels or performs poorly can be analysed, providing a better understanding \new{of} why rather than when algorithms work well or not.
For a conceptual overview of the ISA framework we refer to \citet{Smith-Miles2023}, and to \citet{Kletzander2021,DeCoster2021,Lopes2013,Smith-Miles2011b,Lopes2010} for applications of ISA within the context of personnel scheduling and educational timetabling.

The analysis conducted in this paper is referred to as a `problem type analysis' and diverges from the ISA framework in the sense that features considered in this work are based on the problem type (i.e., number and types of constraints) rather than detailed properties for which the full problem instance needs to be available.
\new{As such, the problem-type analysis presented in this paper is situated at the `higher level' of the problem type (i.e., groups of instances having the same structural properties), whereas the original ISA framework results in a more fine-grained analysis at the level of individual problem instances.}
The choice of features on the level of the problem type is twofold.
The first reason is that practitioners often need to decide what software package to buy or implement, without knowing the full details of the problem instances to be solved in the future.
This rules out the possibility to use detailed features on the level of the problem instances, like the pre-processing time of an associated integer programming (IP) formulation, which is much harder to predict without knowing the precise problem instances.
Similarly, in the context of timetabling, it is reasonable to assume that only one software package can be used to solve all future problem instances. 
While features on the level of problem instances may be quite different, the problem type usually remains relatively constant over time.
The second reason for stepping away from features on the level of problem instances is their lack of intuitiveness, making it hard to derive algorithm insights.
For instance, what fundamental insights can be gained if it is observed that, e.g., a simulated annealing algorithm performs best when the pre-processing of an IP solver is high?
This was also observed by \citet{DeCoster2021}, who restrict the inclusion of instance features to `the most intuitive ones' when using ISA to explain why algorithms perform well.

\begin{figure}
	\centering
	\includegraphics[width=0.6\linewidth]{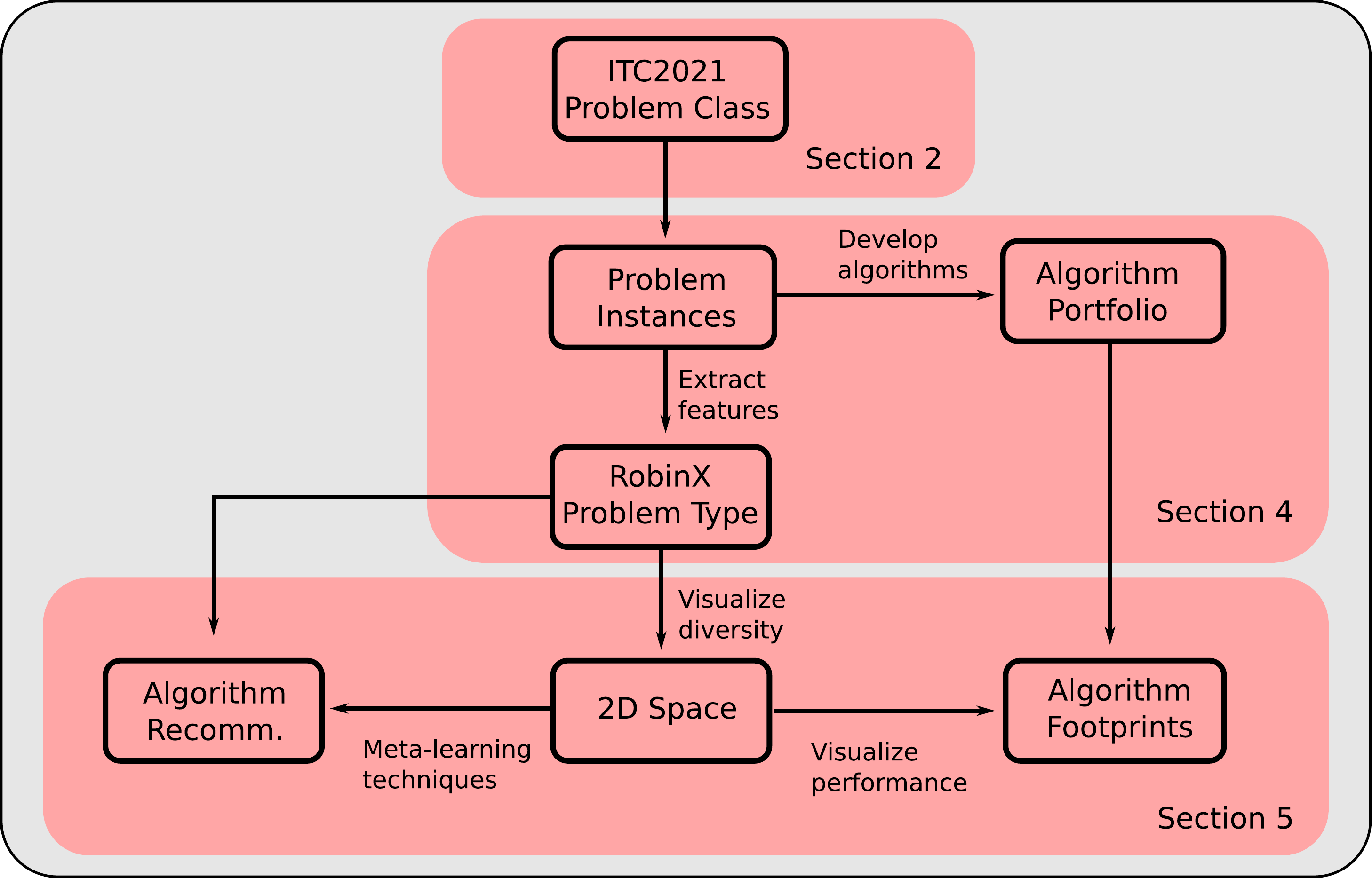}
	\caption{Building blocks of the problem type analysis framework \new{and their link with the sections of this paper}.}
	\label{fig:algorithmSelection}
\end{figure}

The sports timetabling problem analysis in this study extends the work by \citet{VanBulck2022b}, who show how to construct a generator that is able to control for the number of teams and constraints of each type. Given such parametrized generator, they show how to use IP methods to generate a problem instance as close as possible to any given target in the 2D problem type space. This circumvents the need to devise complicated generators that evolve existing problem instances to fill gaps in the 2D space (see, e.g., \citet{Smith-Miles2015}). Although \citet{VanBulck2022b} use their method to come up with a diverse set of 45 problem instances as used in the ITC2021 competition, they do not perform a problem type analysis as they do not show how algorithms perform in this space. Neither do they explain when and why good or bad performance is observed.
For a visual representation of the problem type analysis framework, see \Cref{fig:algorithmSelection}.

\section{Collecting metadata}
\label{sec:metadata}

In this section, we provide an overview of the metadata (i.e., features, problem instances, and algorithms) we have collected to construct the problem type space for the ITC2021 sports timetabling problem.

\subsection{Features}

In order to represent the problem instances on the level of their associated problem type, \citet{VanBulck2022b} propose to use the number of teams in the problem instance as a first feature. A second feature indicates whether or not the competition is phased. Finally, one additional feature per hard and soft constraint type denotes how many constraints of that type are present in the problem instance.
For instance, if a problem instance contains 20 hard constraints of type CA1, then $f_{\text{CA1}}^H = 20$.
Before counting the number of constraints of each type, we first write the constraints in their most elementary form (e.g., splitting a capacity constraint that forbids to play home in multiple slots into a series of constraints each forbidding to play home into one specific slot).
For an overview of the problem type features used in this paper, we refer to \Cref{tab:problemTypeFeatures}.

\begin{table}
	\centering
	\footnotesize
	\ra{1.1}
	\begin{tabular}{l l}
		\toprule
		Name & Description\\
		\midrule
		$f_{|T|}$     & Number of teams\\
		$f_\text{P}$  & Boolean which is one if the tournament is phased and 0 otherwise\\
		$f^H_{\text{CA1}}$   & Number of CA1 hard constraints (others: $f^H_{\text{CA2}}$, $f^H_{\text{CA3}}$, $f^H_{\text{CA4}}$, $f^H_{\text{GA1}}$, $f^H_{\text{BR1}}$, $f^H_{\text{BR2}}$)\\
		$f^S_{\text{CA1}}$   & Number of CA1 soft constraints (others: $f^S_{\text{CA2}}$, $f^S_{\text{CA3}}$, $f^S_{\text{CA4}}$, $f^S_{\text{GA1}}$, $f^S_{\text{BR1}}$, $f^S_{\text{BR2}}$, $f^S_{\text{FA2}}$, $f^S_{\text{SE1}}$)\\
		\bottomrule
	\end{tabular}
	\caption{Overview of problem type features}
	\label{tab:problemTypeFeatures}
\end{table}

In order to quantify the added value of working with features on the level of the problem type rather than more detailed features on the level of the problem instance itself, we also consider the following features referred to as `instance features'.
A first instance-specific set of features is based on model statistics of a straightforward IP formulation proposed by \citet{VanBulck2022b}.
In particular, we report the number of constraints and variables of each type in the model, general statistics like the number of non-zero elements and the mean and standard deviation of the coefficient matrix, right-hand-side and objective coefficient vectors, and statistics on the number of decision variables per constraint and number of constraints in which each variable is involved.
Moreover, after solving the linear-programming (LP) relaxation of this formulation, we count the number of binding constraints, number of fractional primal variables, the variance and the mean of the slack on the constraints, the time to solve the LP-relaxation, the number of simplex iterations needed, and the objective value of the LP-relaxation.
These or very similar features have been used for performance prediction, e.g., in LP (see \citet{Bowly2019}), Mixed-Integer Programming (MIP) (see \citet{Hutter2014}), and combinatorial auctions (see \citet{Leyton-Brown2009}).
A second category of instance features consists of so-called probing features that are obtained by running the simulated annealing algorithm from team Udine (see Section~\ref{subsec:Udine}) with a limited budget of 10,000 evaluations for stage 1 and 2, and 1,000 evaluations for stage 3. This resulted in runtimes in the order of about 20 seconds or less. From this run, we collected the quality of the timetable and number of violated constraints (overall and per constraint type), the runtime and the algorithm's internal cost function (overall and per stage). 
Very similar probing features were used for performance prediction in, e.g., educational timetabling (see \citet{DeCoster2021}) and job shop scheduling (see \citet{Strassl2022}).
This way, a total of 58 features on the level of the problem instances were considered. After removing all features with a Pearson correlation with regard to algorithm performance less than 0.4 (see \Cref{subsec:problemTypeSpace}), a total of 14 instance features remained (see \Cref{tab:instanceFeatures}).

\begin{table}
	\centering
	\footnotesize
	\ra{1.1}
	\begin{tabular}{l l}
		\toprule
		Name & Description\\
		\midrule
		\textbf{IP Model statistics}\\
			\new{$\phi^{\textsc{ip}}_\text{nonzeros}$}                  		 & Number of non-zero elements in the coefficient matrix\\
			\new{$\phi^{\textsc{ip}}_\text{obj std}$}          	 		     & Standard deviation of objective coefficients\\
			\new{$\phi^{\textsc{ip}}_\text{obj mean}$} 		                 & Mean objective coefficient\\
			\new{$\phi^{\textsc{ip}}_\text{cons degree max}$}                  & Maximal number of variables per constraint\\ 
			\new{$\phi^{\textsc{ip}}_\text{cons degree mean}$}     	         & Mean number of variables per constraint\\ 
			\new{$\phi^{\textsc{ip}}_\text{var degree mean}$}      	         & Mean number of constraints in which each variable is involved\\ 
			\new{$\phi^{\textsc{ip}}_\text{obj mean normed}$}      	         & Mean objective coefficient divided by mean left-hand-side coefficient\\ 
			\new{$\phi^{\textsc{ip}}_\text{LP objective}$}	      		     & Value of the LP-relaxation\\[5pt]
		\textbf{Probing Features}\\
			\new{$\phi^\textsc{sa}_\text{CA2}$} 			                & Infeasibility value from violated CA2 hard constraints\\
			\new{$\phi^\textsc{sa}_\text{CA3}$}	 			            & Infeasibility value from violated CA3 hard constraints\\
			\new{$\phi^\textsc{sa}_\text{CA4}$}	 			            & Infeasibility value from violated CA4 hard constraints\\
			\new{$\phi^\textsc{sa}_\text{SA time stage 1 2}$} 	        & Total time needed by the first two stages of the SA algorithm\\
			\new{$\phi^\textsc{sa}_\text{soft cost stage 1 2}$} 		    & Penalty from violated soft constraints after the first two stages of the SA algorithm\\
			\new{$\phi^\textsc{sa}_\text{SA soft cost}$}	 		        & Penalty from violated soft constraints after all three stages of the SA algorithm\\
		\bottomrule
	\end{tabular}
	\caption{Overview of problem instance features}
	\label{tab:instanceFeatures}
\end{table}

\subsection{Problem instances}

\begin{figure}[tp]
	\center
    	\begin{subfigure}[t]{0.45\linewidth}
        	\includegraphics[width=\linewidth]{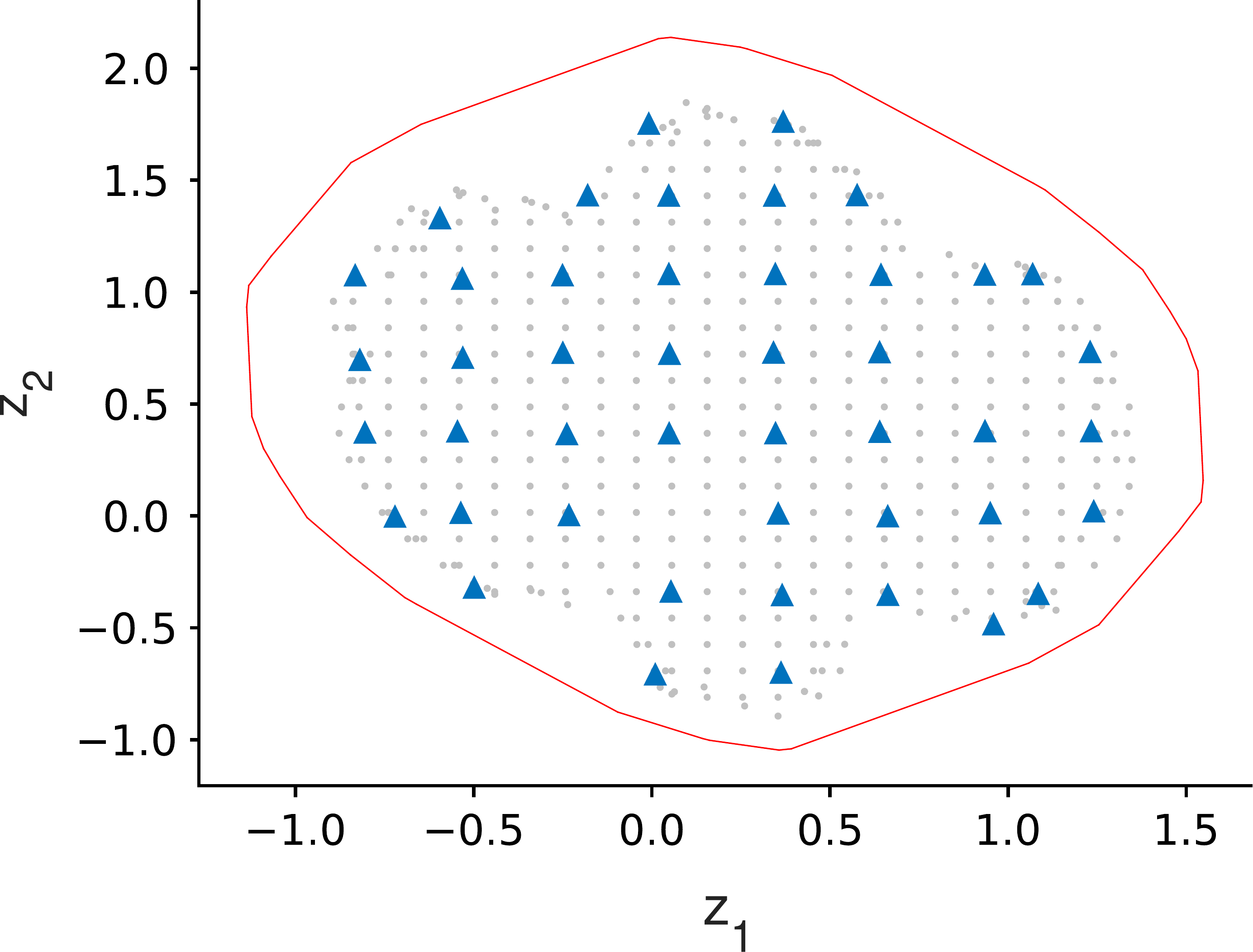}
	\end{subfigure}
	\caption{Projections of the problem instances in the original two-dimensional PCA problem type space. 
    \new{The first and second principal components are denoted by $z_1$ and $z_2$, respectively (see \citet{VanBulck2022b} for the projection matrix used).}
 Grey and blue circles and blue triangles represent the set of training, validation, and ITC2021 problem instances, respectively. }
	\label{fig:itcOriginal}
\end{figure}

The ITC2021 competition offers a benchmark of 45 highly-constrained problem instances, with 16, 18, or 20 teams\footnote{Problem instances can be downloaded from \url{www.itc2021.ugent.be}.}. 
\citet{VanBulck2022b} show how the problem type features from the previous section can be used to represent the ITC2021 problem instances, after applying min-max normalization and mean-centring of the feature values, in a 2D problem type space using principal component analysis (PCA). 
The result is depicted in \Cref{fig:itcOriginal}, where the blue triangles correspond to the ITC2021 instances and the red convex hull to what \citet{VanBulck2022b} call the `target instance space', defining the region in the 2D instance space where real-world problem instances are likely projected. 
\new{This region was determined based on several real-world problem instances presented in the literature.
\citet{VanBulck2022b} also provide a problem instance generator when given the problem type.
This generator was parameterized such that the constraints  (e.g., the minimal separation imposed by an SE1 constraint) represent real-world-like requirements as close as possible.
}

Since a set of 45 problem instances is quite limited to apply machine learning tools, we use the generator and method described by \citet{VanBulck2022b} to produce a diverse set of 518 additional challenging and realistic problem instances that are well scattered in this 2D space (see the grey circles in \Cref{fig:itcOriginal}).
In the remainder of this paper, the set of 45 competition instances is referred to as \emph{ITC2021}, and the set of 518 additional problem instances as \emph{Additional}.
The set of additional problem instances is further split into a set of training (90\%) and validation (10\%) instances.
The combination of the validation and official competition instances is referred to as \emph{Test}.

We remark that the use of PCA for constructing the 2D space was common in the past, but is considered nowadays suboptimal to the approach by \citet{Munoz2018} which explicitly establishes linear trends for feature distribution and algorithm performance.
Notably, the use of this technique assumes the availability of performance results for at least one algorithm, a condition not met at the time of the ITC2021 competition announcement, leading to the use of PCA in that context.
With the availability of algorithm results now, however, the remainder of this paper makes use of the customized projection technique by \citet{Munoz2018}.

\subsection{Sports timetabling algorithms}

This section provides a short description of eight state-of-the-art algorithms that were developed for solving sports timetabling problem instances in the ITC2021 format.
For an overview of the algorithms and their implementation details, we refer to \Cref{tab:algImpl}.

\subsubsection{DES}
The DES algorithm proposed by \citet{Phillips2021} uses an Adaptive Large Neighbourhood Search (ALNS) matheuristic. This can equivalently be described as a fix-and-optimize method (see \Cref{subsub:goal}), with the addition of an adaptive control strategy from the field of reinforcement learning.

To generate a starting solution, DES first attempts to solve a monolithic integer program. If unsuccessful after a fixed time, it reverts to using the canonical factorization from \citet{Werra1981}. The remaining time is spent applying the ALNS technique where only part of the solution may be modified in each iteration. An IP is solved within each neighbourhood subproblem, aiming to minimise the number of hard and soft constraint violations. The neighbourhood definition is adaptively chosen from a set of different types and sizes, based on results of previous iterations. This is treated as a multi-armed bandit problem, using the upper confidence bound method (see \citet{Sutton2018}). Each neighbourhood type (`arm') is chosen as that with the greatest optimistic upper bound on its expected probability of improving the solution (`reward'). This balances between exploring different neighbourhood types and exploiting those which have already performed well. Searching a larger number of small neighbourhoods proved to be the most effective.

\subsubsection{DITUoIArta}
DITUoIArta is a hybrid approach, where an exact solver is combined with a metaheuristic (see \citet{Dimitsas2022b}). The exact solver of choice is the open-source Google ORTools CP/SAT that holds the distinctiveness of expressing a model in Constraint Programming (CP) terms, with the solver reformulating the problem in a satisfiability (SAT) model. The latter seems to be better adapted to the overly constrained nature of sports timetabling. For the local search, simulated annealing is used with five neighbourhood operators proposed in the literature, namely SwapHomes, SwapRounds, SwapTeams, PartialSwapTeams and PartialSwapRounds (see \citet{Anagnostopoulos2006}).

Using CP/SAT, an initial solution that satisfies the base constraints is created. Hard constraints are interpreted as soft constraints, while the original soft constraints are ignored. Simulated annealing then attempts to resolve any remaining violations. If a feasible solution is found, hard constraints become mandatory and soft constraints are turned on. Every time the simulated annealing seems unable to make progress, a CP/SAT improvement process tries to make the present solution better. To do this, a certain number of teams, slots, games, or some combinations of the aforementioned is randomly chosen and kept fixed, while the rest of the current solution is free to change. Finally, the whole process stops if an optimal solution is found either because it has no cost or the solution for the whole problem without fixed parts is reported optimal by the solver. A secondary stopping criterion also exists: if a solution remains unimproved for ten minutes the whole process stops.

Note that in all problem instances there are at most two hard CA3 constraints, one for home games, and one for away games. However, if both constraints appear, then no team can have two breaks in a row. By 
providing this information to the CP solver early and tightening the model, better solutions or feasible solutions for instances that incorporate both hard CA3 constraints were obtained.

\subsubsection{UoS}
The UoS matheuristic approach proposed in \cite{Lamas-Fernandez2021} is based on an IP formulation of the problem. The algorithm starts from a solution that has a valid schedule structure for a (phased) double round robin tournament, obtained with a reduced IP formulation. A Variable Neighbourhood Descent (VND) framework is then used to explore five neighbourhoods that consist of IP models in which some variables are fixed. 
Three of the neighbourhoods look at fixing most of the variables in a current solution, except variables relating to selected sets of $c_1$ slots, $c_2$ teams, and $2$ teams and $\frac{c_1}{2}$ slots, where $c_1$ and $c_2$ are hyperparameters of the algorithm. The two remaining  neighbourhoods optimize one phase of the competition in full and the home team of matches, respectively. To diversify the search at local optima, the objective function coefficients of most violated constraints increase, prompting the algorithm to move to other parts of the search space. The full IP formulation of the problem, and more details of the algorithm and its neighbourhoods are reported in \cite{Lamas-Fernandez2021}.

For this article, the algorithm was divided into two phases: feasibility search, and improvement search. The feasibility search phase tackled the problem discarding all soft constraints but aiming to minimise breaks, and applied the VND matheuristic for $25\%$ of the time. During this phase, models were solved only for up to two seconds. If a feasible solution was found, the algorithm moved to the improvement phase, otherwise the feasibility phase was extended until a feasible solution was found with ten seconds. The improvement phase applied VND with the full model for the remaining time. For all instances, five different 48-hour runs were performed, with slight variations of hyperparameters. Four runs were set up with $c_1 = 12$, $c_2 = 8$ and time limits from 10 to 60 seconds. The fifth run considered smaller models with $c_1 = 10$, $c_2 = 6$ and a 15 second time limit.

Finally, a further 24-hour run was set up for those instances that remained infeasible after the above procedure. Here, the algorithm used in the feasibility search was slightly different, with the intention of tackling larger problems. In this version, the problems are decomposed in stages, each stage considering 200 constraints more than the previous stage. Only when all the constraints of a stage are satisfied, the next 200 are added to the model, prioritised so that the most violated constraints are added first.
Further, the selection of slot sets in our neighbourhoods included evaluating a few options (5 in our experiments) by means of the lower bound provided by their linear relaxation. The algorithm then selects the set in which this value is lower. If none of the values improves the current best solution, more sets are generated, potentially modifying the number of slots in the set (effectively adapting the value of $c_1$).

\subsubsection{Goal}
\label{subsub:goal}
The Goal algorithm proposed by \cite{Fonseca2022} is based on the fix-and-optimize strategy, which may be categorized as a matheuristic. 
Matheuristics are heuristics that take advantage of the power of mathematical programming solvers to tackle hard combinatorial optimization problems. 
More specifically, fix-and-optimize algorithms are matheuristics that iteratively employ a mathematical programming solver to optimize a small subproblem while the remainder of the problem remains fixed.

Goal selects, at each iteration, a subset of the time slots to compose the subproblem. Their related variables are released while the others are fixed to their current values. Variables related to the inversed venue games are also released to allow venue exchange. Hopefully, in a given iteration, a better solution will be found by the IP solver by exploring the possibility of exchanging several variables simultaneously and efficiently. The subproblem size is auto-adjusted throughout the execution of the algorithm, while the time limit for each iteration was empirically set to 100 seconds. 

Fix-and-optimize algorithms require an initial solution to heuristically improve, which, in this case, is based on the canonical factorization from \citet{Werra1981}.
Given this initial solution, the proposed algorithm executes twice: first to obtain a feasible
solution and then to refine (improve) this solution.
Only hard constraints, modelled as soft constraints, are included in the first execution.
When a feasible solution (zero-cost) is found, the second execution begins, taking this feasible solution as input. In the second execution, all slack variables for hard constraints are removed and the soft constraints are added to the model. 

The considered stopping criterion is a runtime limit of 24 hours. The algorithm may also be interrupted when optimality can be asserted - a zero-cost solution has been found or the subproblem size matches the problem size and the subproblem solution is optimal. 

\subsubsection{FBHS}
Instead of constructing a timetable all at once, the first-break-heuristically-schedule (FBHS) algorithm proposed by \citet{VanBulck2022c} considers the following subproblems which are sequentially solved:
\begin{paraenum}
    \item determine for each team in every time slot whether it plays home or away (i.e., its HAP), and
    \item determine the opponent of each team in every time slot (opponent schedule).
\end{paraenum}
These two steps, though, must be compatible: two teams can only play against each other in the opponent schedule when one team plays home and the other team plays away according to their HAPs. It is important to realize that a HAP set does not necessarily allow a (high-quality) compatible opponent schedule that schedules all required matches (see, e.g., \cite{Briskorn2008b}). Hence, to generate the HAP set, we use a traditional Benders' decomposition approach that enforces some necessary HAP set feasibility conditions while also considering the LP-relaxation of the best possible opponent schedule.

The time limit for the HAP set generation process was set to 6 hours (wall time).
Given the most promising HAP set found in step 1, a compatible opponent schedule is constructed using a fix-and-optimize matheuristic (a time limit of two minutes per IP model was imposed).
The matheuristic was repeated for a total of 24 times. 
On average, the compatible opponent schedule generation took 13.5 hours (a total time limit of 24 hours per instance was enforced).

\subsubsection{Udine}\label{subsec:Udine}

The Multi-Neighbourhood Search proposed by \cite{Rosati2022} employs a portfolio of six local search neighbourhoods\footnote{The source code is available at: \mbox{\url{https://github.com/robertomrosati/sa4stt}}.
}. Five neighbourhoods (SwapHomes, SwapTeams, SwapRounds, PartialSwapTeams, PartialSwapRound) were previously proposed in the literature \cite{Anagnostopoulos2006}, while the last one (PartialSwapTeamsPhased) is novel and relies on the concept of \emph{mixed phase}, by performing partial swaps among the opponents of two teams in one of the two \emph{mixed legs}. In this way, the move does not add violations to the 
phase constraint, possibly requiring additional swaps of the home advantage. PartialSwapTeamsPhased proved its effectiveness especially with regard to phased instances, but it is useful, according to experimental data, also on non-phased ones.


The metaheuristic that guides the search is simulated annealing, with the addition of the so-called cut-off mechanism to speed up the early phases of the search (see \citet{Johnson1989}).
The search is performed in three sequential simulated annealing runs, called stages, that are aimed, respectively, at finding a feasible solution that does not violate hard constraints, at exploration and exploitation by navigating both the feasible and the infeasible region, and at finding a better local minimum by just exploring the feasible region. The first stage starts from a greedily-constructed solution, every next stage is warm-started by the best solution found in the previous one. In line with \citet{Rosati2022}, the stages always execute a given number of local search evaluations, so there is no fixed time limit. The resulting running times vary noticeably, as the time spent for the evaluation of local search moves depends on the size of the instance and on the number of constraints. 


\subsubsection{Reprobate}

Reprobate generates a monolithic encoding of the timetabling problem as a weighted pseudoboolean (PB) constraint optimization problem and applies a portfolio of off-the-shelf PB solvers (see \citet{Lester2022a}). PB constraints are a generalisation of boolean satisfiability (SAT) constraints over boolean variables. Many relationships involving finite sets, such as set membership, subset and set cardinality, have a succinct encoding in this format, which makes it well-suited to a range of discrete optimization problems. Furthermore, PB constraints are equivalent to 0-1 integer programming. However, compared to MIP solvers, PB solvers tend to be stronger on problems solvable using mainly boolean reasoning, but weaker on problems where heavy use of arithmetic reasoning is required. 
Reprobate also tries some variations on the default encoding and applies a single solver to those.


If a feasible solution is found, and somewhat in line with the well-known paradigm of first-schedule-then-break (see \citet{Trick2001}), Reprobate attempts to improve the objective
by encoding and solving a second optimization problem in which
the allocation of opponents in each slot is fixed, but which team plays home in each game is not, effectively allowing the home away pattern (HAP), i.e\ a team's succession of home and away matches, to be tuned.
Again, this is a PB constraint problem.

For this study, the portfolio of solvers used in \emph{Reprobate} was updated. Reprobate now uses version 3.3.9 of \emph{clasp} (running with the \emph{crafty} preset, see \citet{Gebser2007}) as the default solver. 
\emph{Sat4J} remains in the portfolio, but only with the specialist PB algorithms introduced in version 2.3.6  (see \citet{LeBerre2021}); the default setting is no longer used, as its performance was poor.
In its place, the latest version of \emph{RoundingSat} was used,
with \emph{toyconvert} to translate between the WBO format (produced by \emph{Reprobate}) and OPB format (consumed by \emph{RoundingSat}, see \citet{Elffers2018}).
\emph{RoundingSat} was compiled without the optional \emph{Soplex} integration,
as it seemed to degrade performance on the benchmarks;
this also means that all the code and solvers remain available under permissive open source licences.

Each solver instance was run with a time limit of 10 minutes. With 6 solvers on the default encoding, 7 encoding variations with the default solver and up to 2 tuning phases, this gives a total time limit of 150 minutes. In practice, it was rare to exceed 140 minutes, as one encoding variation (ignoring all soft constraints) often led to a solution in a few seconds, and when it did not, there was often no feasible solution to tune.

\subsubsection{MODAL}
\citet{Berthold2021} model the problem as a MIP and propose a solution strategy, coined MODAL, based on an ensemble approach employing different MIP solvers (Gurobi, CPLEX, Xpress). 
The choice for an ensemble approach is twofold.
Firstly, as MIP solvers have different strengths, some solvers may be more suitable than others depending on the problem instance to be solved.
The second reason is to exploit the effect of performance variability: instead of computing solutions in single long runs, the solvers are called repeatedly, starting with the best found solution in each iteration. 

As finding an initial solution can be challenging, two start procedures have been implemented. The first one uses an analytic centre objective. It replaces the objective function of a MIP by coefficients that correspond to the analytic centre of the polyhedron associated to the MIP. The second uses a version of the feasibility pump with multiple integer reference vectors (see, e.g., \cite{Berthold2019}). 

Using a dedicated cluster facility, MODAL was run for each instance up to 25 times, using an overall time limit of up to 10 hours per instance.
The parameters were varied between the runs, usually increasing the number of heuristic runs, emphasizing on finding feasible solutions, reducing the amount of cutting plane generation, etc. 
Note, though, that all instances went through the same loop of repeated solving by various solvers, with the amount of loops depending on the solution quality. Similarly, the use of heuristic methods to ignite the search depends on whether the initial MIP searches yield a feasible solution, but not on any of the problem characteristics.

The catch of MODAL is that it does not use any special purpose timetabling algorithm. All methodology is general MIP technology, available independently of this work.


\begin{singlespace}
\begin{table}
\footnotesize
\centering
	\ra{1.2}
     \begin{tabular}{l p{58pt} p{100pt} p{150pt} p{57pt}}
    \toprule
    Algorithm & Search method & Software details & Hardware details & Clock speed ratio\\
    \midrule
    DES         & Adaptive LNS matheuristic     & Python 3.10, Gurobi 10.0 & Intel Xeon with 4 cores and 8 threads (Google Compute Engine ``c2-standard-8'') for 2.5 hours per instance &3.9/3.9\\
    DITUoIArta  & CP/SAT + Simulated annealing      & Python 3.10, ORTools 9.4  & 6 x Intel Core i5-10505  with 8GB RAM (all cores activated for the solver only) for 1 hour each time for each instance (an instance may run multiple times though) & 3.2/3.9\\ 
    UoS         & VND matheuristic      & Python 3.10.4, Gurobi 9.0.2 & Dual Intel Skylake with 4 or 20 cores &2.0/3.9\\
    Goal        & Fix-and-optimize matheuristic & Java 16, Gurobi 10.0 & Intel Core i7-8700 with 12 GB RAM (single thread running for 24 hours per instance) & 3.2/3.9\\
    FBHS        & IP Decomposition + matheuristic  & C++, CPLEX 12.10 &Intel Xeon E5-2660v3 (Haswell-EP) processor enabled with 8 cores &2.6/3.9\\
    Udine       & Multi-Neighbourhood Search   & C++17 & Intel Xeon Processor (Cascadelake), 16 cores, (max one core per execution) &2.4/3.9\\
    Reprobate         & Pseudoboolean optimization & Perl, clasp 3.3.9, Sat4J 2.3.6, RoundingSat Git Nov 2022      & Intel Core i7-8700 with 64 GB RAM (single core running for 2.5 hours per instance) & 3.2/3.9\\
    MODAL       & IP Branch \& Cut & Python, Zimpl, C, Gurobi 10, Xpress & Per instance one thread was used and several instances were run at the same time on a machine with a 24 core / 48 hyperthread Intel Xeon Gold 6342 CPU & 2.8/3.9\\
    \bottomrule
    \end{tabular}
        \caption{Overview of algorithms together with software and hardware details. The last column compares processor clock speeds relative to the `fastest' processor used (3.9 GHz).
    }
    \label{tab:algImpl}
\end{table}
\end{singlespace}


\subsection{Performance criteria}
\label{subsec:criteria}

All algorithms described in the previous section were given two weeks of time to produce best found solutions for all additional problem instances. The algorithms, which in most cases differ somewhat from their ITC2021 implementation, were also rerun on the official ITC2021 competition instances. 
Besides this overall two-week time limit, there were no limits on the computation time for individual problem instances, nor on the number of cores or computing infrastructure used. 
The motivation for this set-up is as follows.
First of all, computation time is rarely a critical issue in sports timetabling, as practitioners typically have several days or even weeks to produce the fixtures, and this has to be done only once per season.
In such a setting, the quality of the timetable is much more important than the computational resources used.
Hence, we allowed participants to run their algorithm as long as they preferred themselves so that their algorithms could run at full potential.
However, the overall two week time limit in combination with the large number of instances was used to prevent excessive running times.

We note that this set-up is different from most other applications in ISA, where the assumption is that a single runtime is employed for all algorithms.
This assumption usually aligns with the central idea that only one algorithm can practically be used to solve a given problem instance, given time or resource constraints that prohibit running algorithms sequentially or in parallel. 
In contrast, our approach diverges from other applications in the literature in that it is not the time limit preventing sequential or parallel algorithm runs; rather, it's the assumption that practitioners intend to acquire or implement only one algorithm from the existing literature.
Even though practitioners are practically indifferent with regard to runtimes, we consider the distribution of runtimes over the problem type space as a secondary but subordinate objective, as we believe an analysis of runtimes may result in additional algorithm insights.

In summary, we consider the following performance criteria to evaluate the performance of the algorithms, where computation time is deemed the least important metric.
\begin{enumerate}
	\item \textbf{Feasible solutions.} As finding a feasible solution is a significant challenge for algorithms for many problem instances, the first metric is the percentage of problem instances for which an algorithm found a feasible solution.
	\item \textbf{Good and best solutions.} While finding a feasible solution is obviously important, so is the quality of the solution found. 
    Therefore, the second metric calculates the percentage of problem instances for which an algorithm identified either the optimal solution or, alternatively, a `good' solution, defined as within 5\% of the best one. 
	\item \textbf{Relative gap.} As recommending a near-optimal algorithm is still better than recommending an algorithm that results in a poor solution, a third measure computes the average relative distance to the best solution over all instances for which an algorithm found a feasible solution.
	\item \textbf{Computation time.} The final performance metric is the total CPU time, normalized for clock speed by multiplying by the clock speed ratio from \Cref{tab:algImpl}, used by the algorithm. 
\end{enumerate}

\section{A Problem Type Analysis for ITC2021}
\label{sec:problemTypeAnalysis}

Given the metadata from the previous section, this section uses the open-source MATLAB software `ISA toolkit'\footnote{The ISA toolkit is available from \url{https://github.com/andremun/InstanceSpace/}} to visualize the problem type space for the ITC2021 sports timetabling problem\footnote{All metadata files containing the instances, associated feature values, and per-instance algorithm results are available from \url{https://robinxval.ugent.be/Docs/MetaData_ITC2021.zip}}.

\subsection{Problem type space}
\label{subsec:problemTypeSpace}

In line with \citet{VanBulck2022b}, all features were first mean-centred and min-max normalized.
Preprocessed features were then offered to the ISA toolkit, which applied an automated feature subset selection, reducing the number of features from 18 to 12. The ISA toolkit does this in order to obtain a subset of features that are uncorrelated with each other but strongly correlated with algorithm performance (see \citet{Smith-Miles2023}).
The remaining 12 features in combination with all training problem instances were then used by the toolkit to construct a linear transformation from the high-dimensional problem type space to a space with two dimensions only. The full projection matrix, making use of the customized projection technique by \citet{Munoz2018}, is given in \Cref{eq:weights}.


\begin{singlespace}
{
	\small
\begin{equation}
	\label{eq:weights}
	\bm{(z_1,z_2)} = 
	\begin{bmatrix} 
		-0.0859	& 	\phantom{-}0.3822  \\[2pt]
		-0.3676	& 	-0.5381 \\[2pt]
		-0.4103	& 	-0.2229 \\[2pt]
		\phantom{-}0.4221	& 	-0.1775 \\[2pt]
		\phantom{-}0.4957	& 	-0.1841 \\[2pt]
		\phantom{-}0.2012	& 	-0.6936 \\[2pt]
		-0.3357	& 	\phantom{-}0.0449  \\[2pt]
		\phantom{-}0.0908	& 	\phantom{-}0.1941  \\[2pt]
		\phantom{-}0.4566	& 	-0.9567 \\[2pt]
		\phantom{-}0.0404	& 	\phantom{-}0.208   \\[2pt]
		\phantom{-}0.2266	& 	\phantom{-}0.2159  \\[2pt]
		-0.3634	& 	-0.2149 \\
	\end{bmatrix}^{\intercal}
	\begin{bmatrix}
		f_{|T|}\\[1.5pt]
		f_{P}\\[1.5pt]
		f^S_{\text{CA1}}\\[2pt]
		f^H_{\text{CA2}}\\[2pt]
		f^H_{\text{CA3}}\\[2pt]
		f^H_{\text{CA4}}\\[2pt]
		f^H_{\text{GA1}}\\[2pt]
		f^S_{\text{GA1}}\\[2pt]
		f^H_{\text{BR2}}\\[2pt]
		f^S_{\text{BR2}}\\[2pt]
		f^S_{\text{FA2}}\\[2pt]
		f^S_{\text{SE1}}\\
	\end{bmatrix}
\end{equation}
}
\end{singlespace}

\begin{figure}[tp]
	\centering
    	\begin{subfigure}[t]{0.4\linewidth}
        	\includegraphics[width=\linewidth]{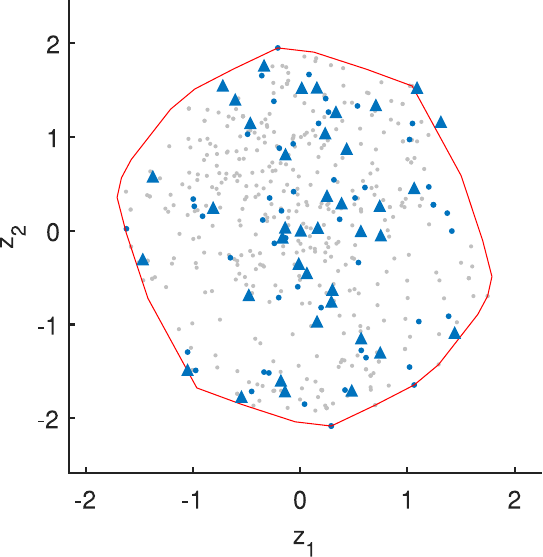}
	\end{subfigure}
	\caption{Projections of the problem instances in the newly generated problem type space. 
\new{The first and second principal components are denoted by $z_1$ and $z_2$, respectively (see \Cref{eq:weights} for the projection matrix used).}
 Grey and blue circles and blue triangles represent the set of training, validation, and ITC2021 problem instances, respectively. }
	\label{fig:itcTrans}
\end{figure}

\Cref{fig:itcTrans} shows the resulting distribution of all instances over the newly generated problem type space.
First of all, it can be seen that there is no clear distinction between the training and test instances (note that the projection matrix is only based on the training instances).
Moreover, even in this newly generated problem type space, it is clear that the set of problem instances can be considered as `diverse' as there are no clear gaps in the 2D space.
If this would not have been the case, we remark that the approach outlined in \citet{VanBulck2022b} can be used to generate problem instances that are as close as possible to any given target coordinates in the newly generated 2D problem type space.

\begin{figure*}
    \centering
    \begin{subfigure}[t]{0.32\linewidth}
        \centering
        \includegraphics[width=\linewidth]{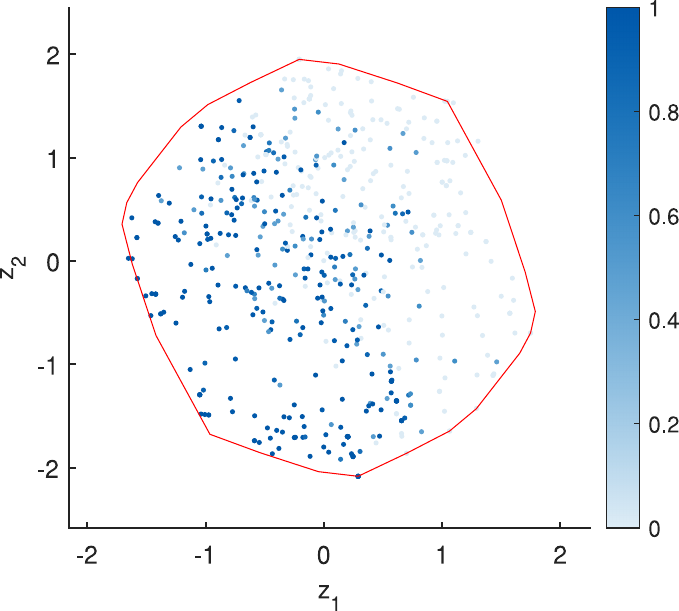}
	\caption{CA1 Soft}
	\label{fig:ca1Soft}
    \end{subfigure}%
    \hfill
    \begin{subfigure}[t]{0.32\linewidth}
        \centering
        \includegraphics[width=\linewidth]{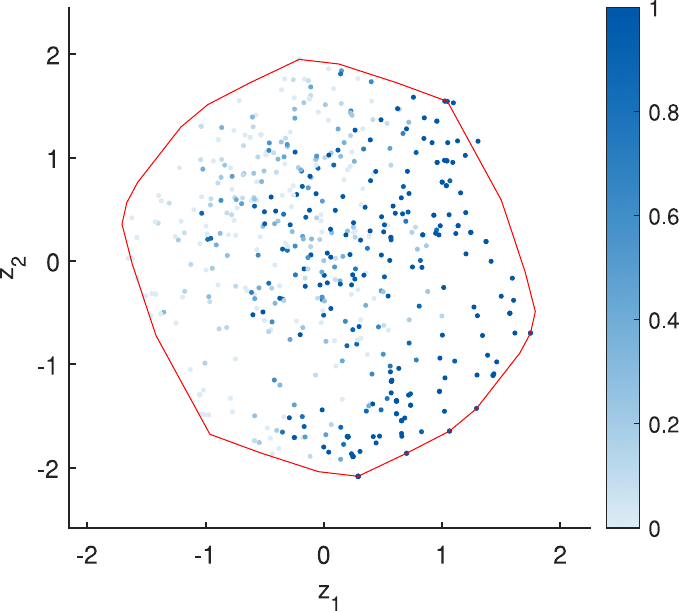}
        \caption{CA2 Hard}
	\label{fig:ca2}
    \end{subfigure}%
    \hfill
    \begin{subfigure}[t]{0.32\textwidth}
        \centering
        \includegraphics[width=\linewidth]{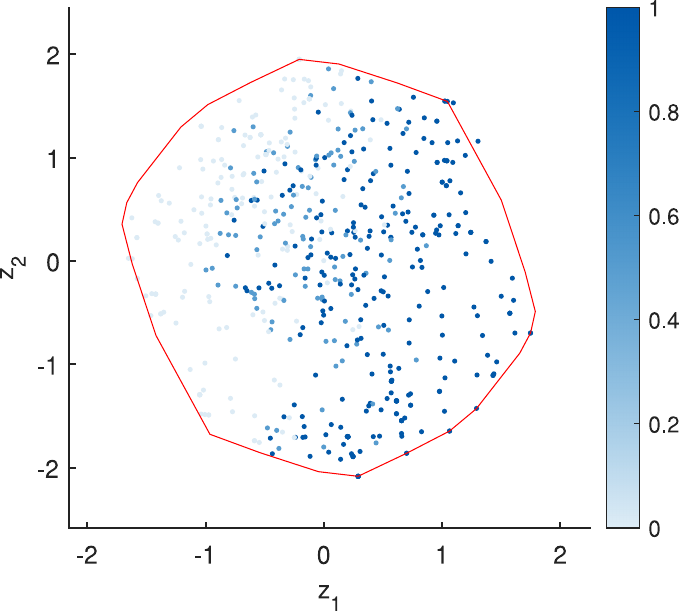}
        \caption{CA3 Hard}
	\label{fig:ca3}
    \end{subfigure}%
    \\
    \begin{subfigure}[t]{0.32\textwidth}
        \centering
        \includegraphics[width=\linewidth]{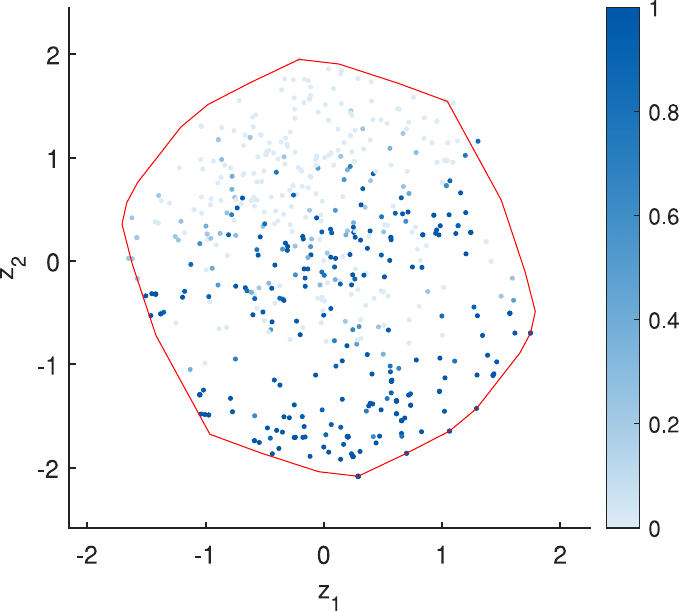}
        \caption{CA4 Hard}
	\label{fig:ca3Soft}
    \end{subfigure}%
    \hfill
    \begin{subfigure}[t]{0.32\textwidth}
        \centering
        \includegraphics[width=\linewidth]{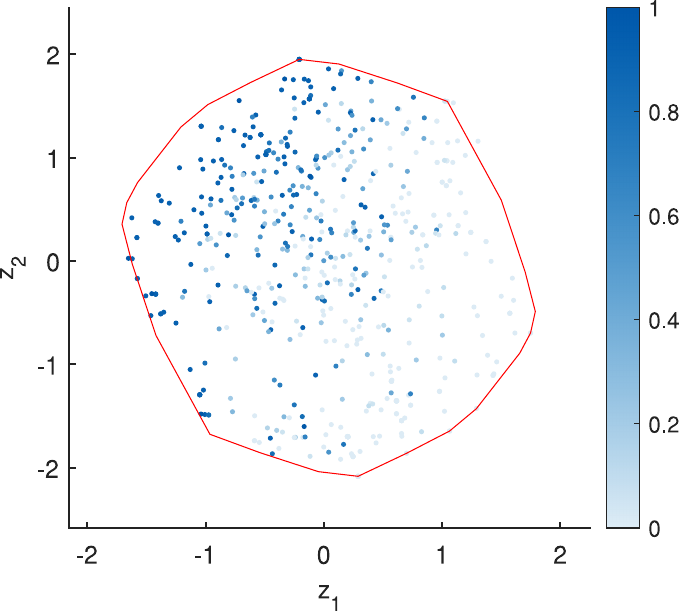}
        \caption{GA1 Hard}
	\label{fig:ca4Soft}
    \end{subfigure}%
    \hfill
    \begin{subfigure}[t]{0.32\textwidth}
        \centering
        \includegraphics[width=\linewidth]{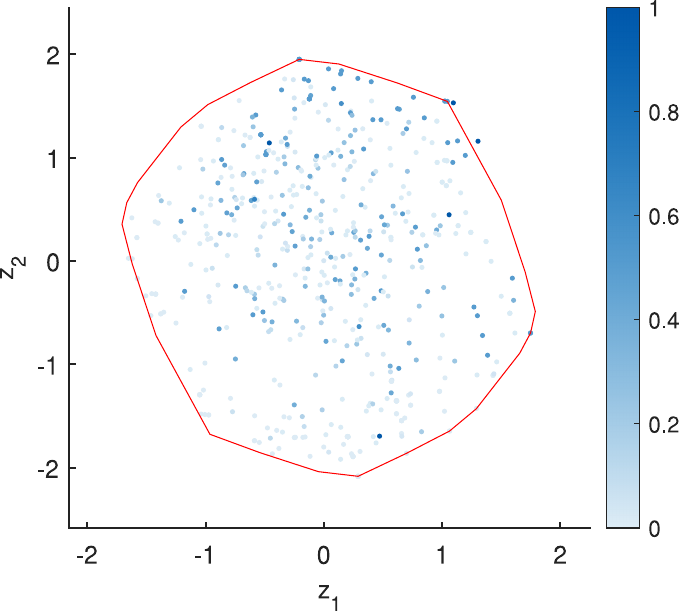}
        \caption{GA1 Soft}
	\label{fig:ga1}
    \end{subfigure}%
    \\
    \begin{subfigure}[t]{0.32\textwidth}
        \centering
        \includegraphics[width=\linewidth]{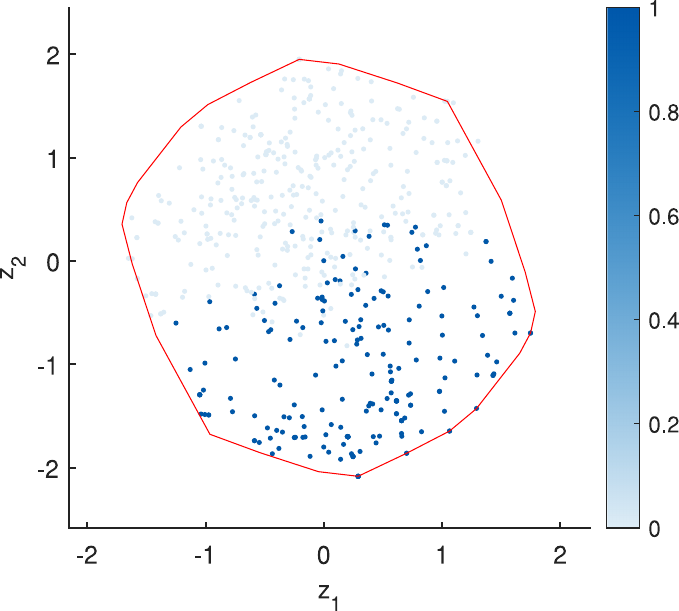}
        \caption{BR2 Hard}
	\label{fig:br2}
    \end{subfigure}%
    \hfill
    \begin{subfigure}[t]{0.32\textwidth}
        \centering
        \includegraphics[width=\linewidth]{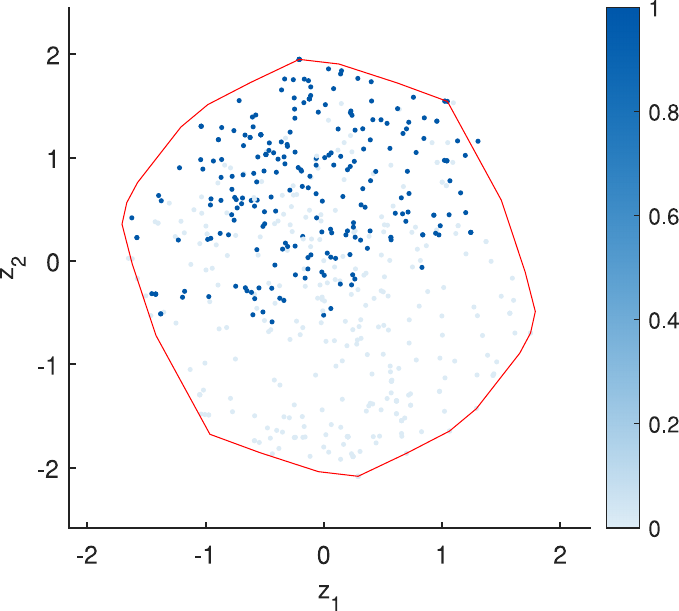}
        \caption{BR2 Soft}
	\label{fig:br2Soft}
    \end{subfigure}%
    \hfill
    \begin{subfigure}[t]{0.32\textwidth}
        \centering
        \includegraphics[width=\linewidth]{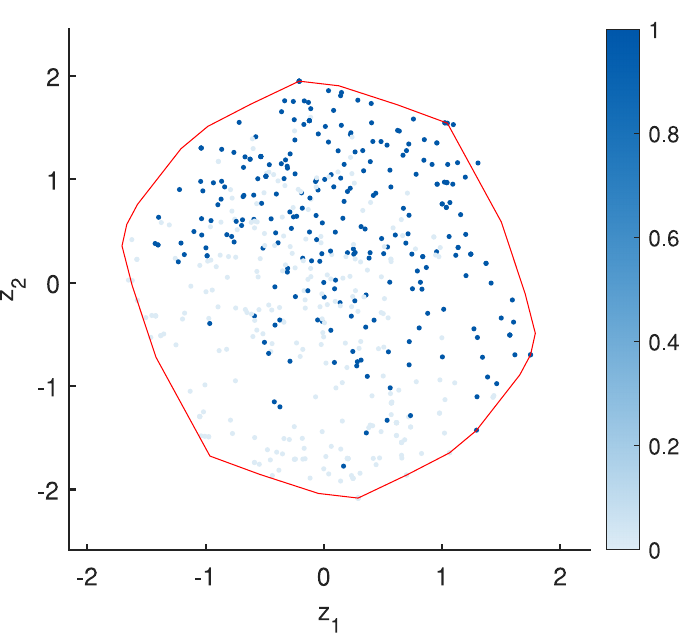}
        \caption{\new{FA2 Soft}}
	\label{fig:ca4}
    \end{subfigure}%
    \\
    \begin{subfigure}[t]{0.32\textwidth}
        \centering
        \includegraphics[width=\linewidth]{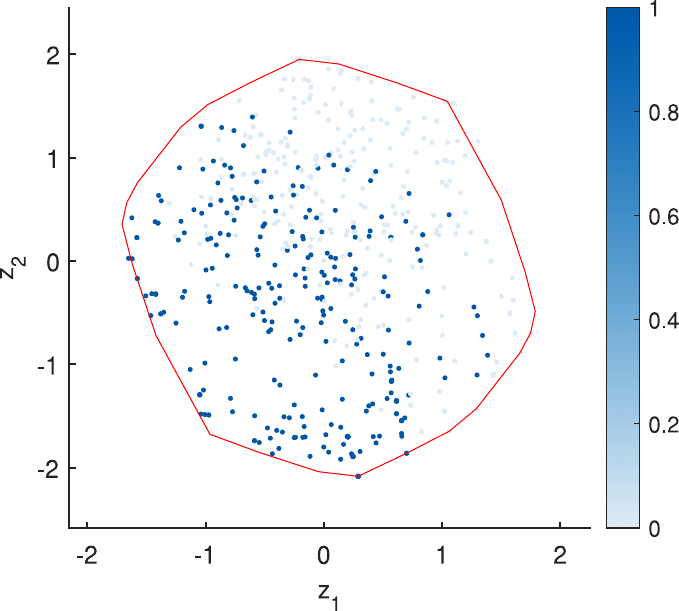}
        \caption{SE1 Soft}
	\label{fig:br1}
    \end{subfigure}%
    \hfill
    \begin{subfigure}[t]{0.32\textwidth}
        \centering
        \includegraphics[width=\linewidth]{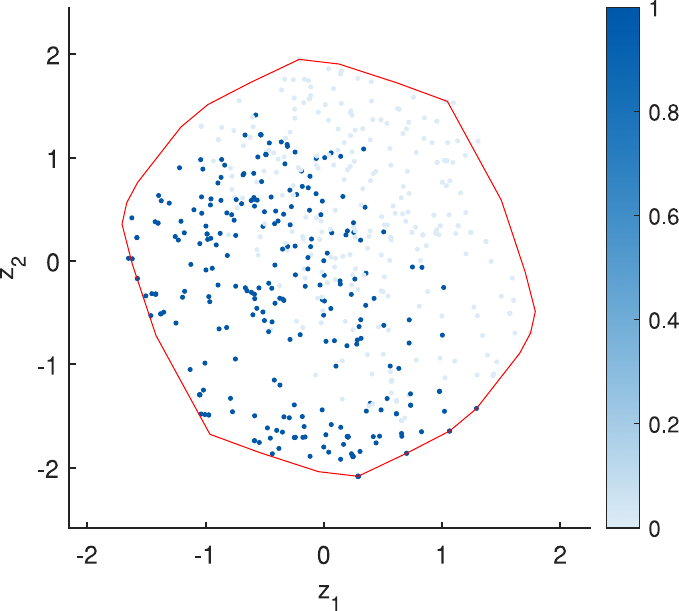}
	\caption{Phased}
	\label{fig:ga1Soft}
    \end{subfigure}%
    \hfill
    \begin{subfigure}[t]{0.32\textwidth}
        \centering
        \includegraphics[width=\linewidth]{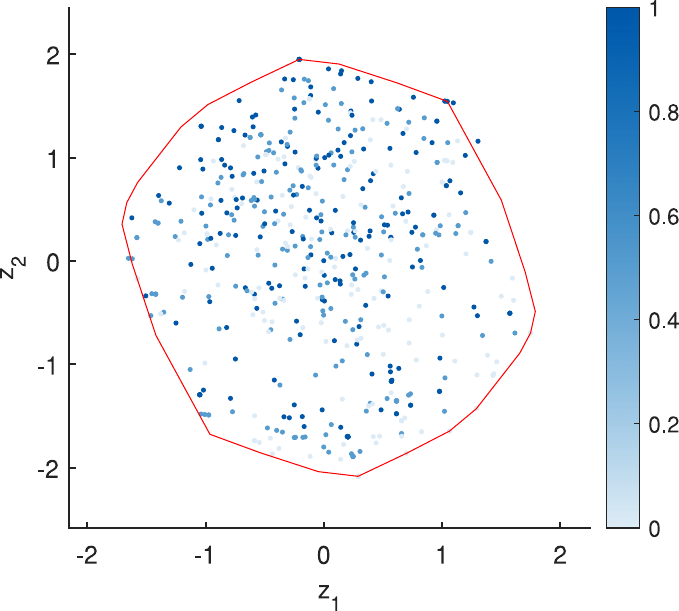}
        \caption{Number of teams}
	\label{fig:br1Soft}
    \end{subfigure}%
    \caption{Normalized feature values across the problem type space \new{as shown by the colorbar on the right of each subfigure.}}
    \label{fig:instanceFeatures}
\end{figure*}

In order to gain insights into the composition of the problem type space, \Cref{fig:instanceFeatures} shows the distribution of the feature values over the training and test set.
As a result of the customized projection method in the ISA toolkit, we observe that the features follow a noticeable linear trend over the problem type space.
For instance, the problem instances with a high value for the  BR2 hard constraint are located near the bottom of the 2D space whereas those with a high value for the BR2 soft constraint are on the top of the 2D space.
Similarly, phased problem instances are located in the left of the 2D space whereas regular problem instances are located near the right.
For the number of teams, however, no obvious linear trend can be observed (this was on purpose, as explained in \citet{VanBulck2022b}).

\subsection{\new{Algorithm performance evaluation}}

\new{This section evaluates the performance of the algorithms based on each of the performance criteria introduced in \Cref{subsec:criteria}.}

\subsubsection{Feasible and best solutions found}

\begin{figure}
	\hfill
    \begin{subfigure}[t]{0.39\textwidth}
	\includegraphics[width=0.95\linewidth]{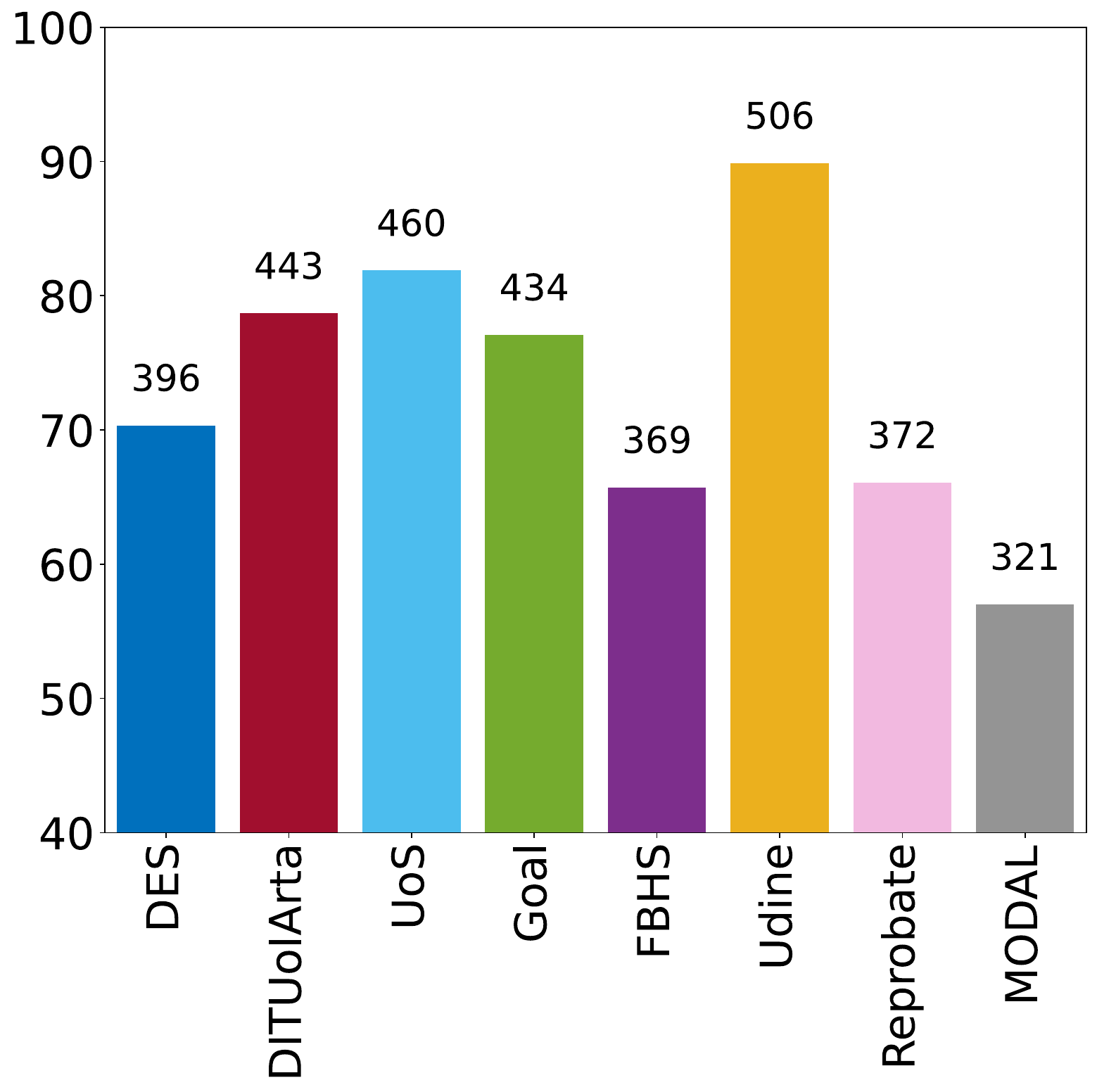}
	\caption{Feasible solutions}
	\label{fig:numberFeas}
    \end{subfigure}
	\hspace{20pt}
    \begin{subfigure}[t]{0.55\textwidth}
	\includegraphics[width=0.95\linewidth]{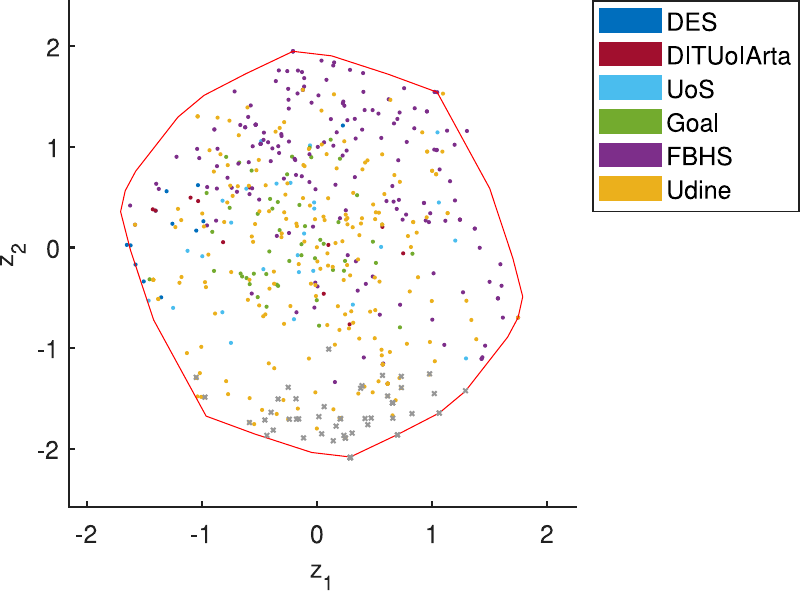}
	\caption{Best solutions} 
	\label{fig:numberBest}
    \end{subfigure}
    \caption{Percentage of feasible solutions found (\textit{a}; numbers on top of the bars denote the absolute number of solutions found), and best performing algorithm for each problem instance (\textit{b}; ties are broken by the ISA toolkit at random).
    \new{Reprobate and MODAL never resulted in a (unique) best solution, and thus do not appear in the figure in the right.}
    }
	\label{fig:numberFeasBest}
\end{figure}

In order to evaluate the performance of the algorithms over the problem instances, \Cref{fig:numberFeas} starts by depicting the number of problem instances for which each of the algorithms found a feasible solution.
Udine managed to find a solution for the largest number of instances, followed by UoS, DITUoIArta, and Goal. 
On the other hand, Modal, FBHS and Reprobate struggle to find feasible solutions for many of the problem instances.
In case of the FBHS solver, however, if a solution is found it is very often a good one as is indicated in \Cref{fig:numberBest} which shows the winning algorithm in terms of the best solution found for each of the problem instances.
\Cref{fig:numberBest} also shows that most of the algorithms perform particularly well in a specific part of the problem type space (which is a first indication of the potential of algorithm selection).
For example, we see that the FBHS solver performs exceptionally well near the top right of the problem type space where the value for the BR2 soft constraints is high (see \Cref{fig:instanceFeatures}).
In contrast, Goal and UoS score well near the origin of the space where, as a result of the mean centring, instances have more or less average feature values.

\begin{figure}
	\hfill
    	\begin{subfigure}[t]{0.45\textwidth}
		\includegraphics[width=\linewidth]{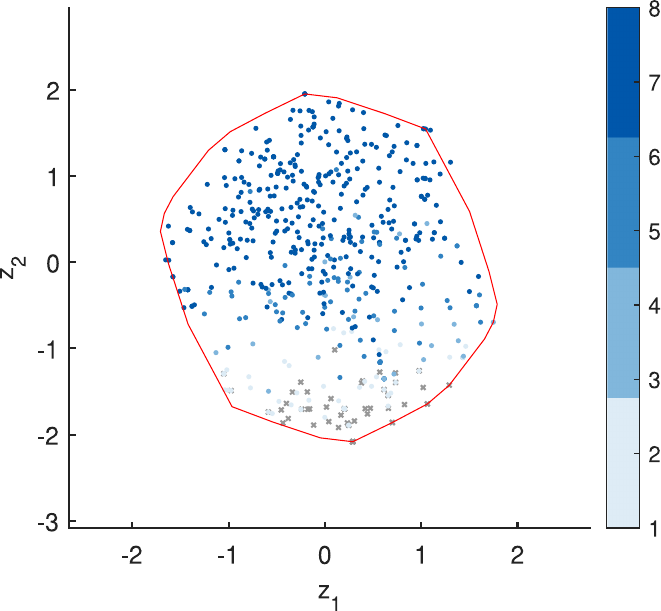}
		\caption{Feasible solutions}
		\label{fig:noFeas}
	\end{subfigure}
	\hfill
    	\begin{subfigure}[t]{0.45\textwidth}
		\includegraphics[width=\linewidth]{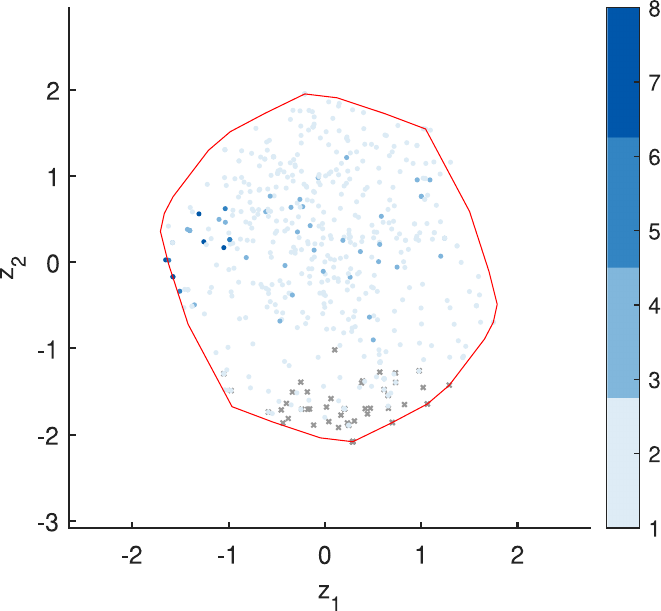}
		\caption{Good solutions}
		\label{fig:noGood}
	\end{subfigure}
	\hfill
	\,
	\caption{Number of algorithms that found a feasible or good solution for each of the problem instances. Problem instances for which no solutions are found are indicated by grey x-marks.}
	\label{fig:instanceHardness}
\end{figure}

In order to locate where the hard and easy problem instances are located, \Cref{fig:noFeas,fig:noGood} shows for each instance the number of algorithms that found a feasible solution and those that found a good solution.
For 51 out of the 518 additional instances and 4 out of the 45 ITC2021 competition instances, no solver found a feasible solution (by design, a feasible solution exists).
From this figure, it is clear that the hard problem instances are located near the bottom of the problem type space.
This part of the problem type space corresponds to problems that are phased or have SE1 soft constraints, in combination with many BR2 and CA4 hard constraints.
On the other hand, near the middle of the  space, and especially near the middle left, there are several instances for which multiple or even all algorithms find a good solution.
Based on \Cref{fig:instanceFeatures}, this area is characterized by the lack of BR2 hard and soft constraints.

\subsubsection{Relative gap}
\begin{figure*}
    \centering
    \begin{subfigure}[t]{0.32\textwidth}
        \centering
        \includegraphics[width=\linewidth]{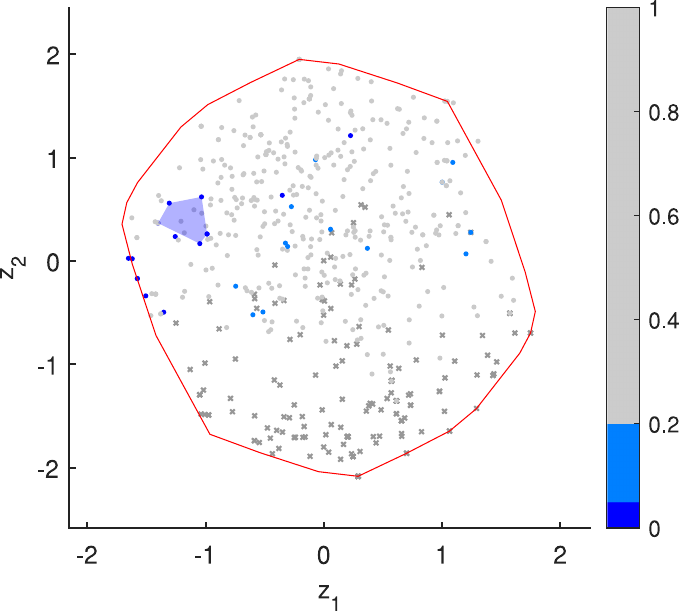}
	\caption{DES}
	\label{fig:regret_des}
    \end{subfigure}
    \hfill
    \begin{subfigure}[t]{0.32\textwidth}
        \centering
        \includegraphics[width=\linewidth]{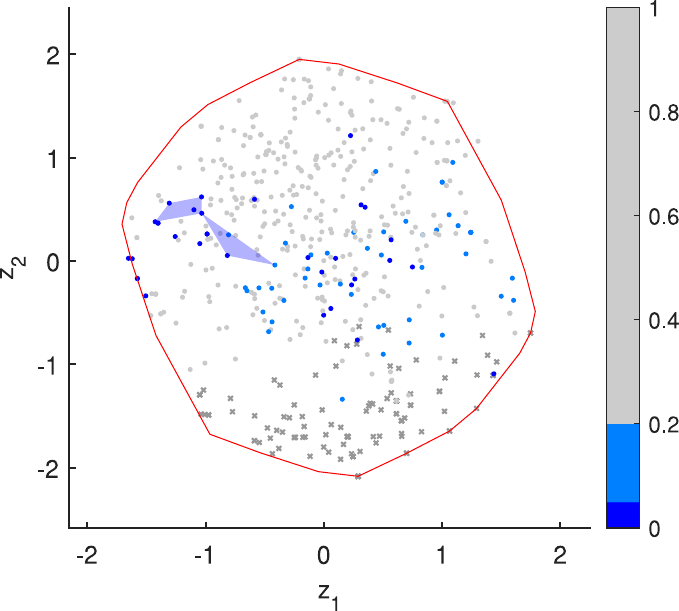}
	\caption{DITUoIArta}
	\label{fig:regret_arta}
    \end{subfigure}
    \hfill
    \begin{subfigure}[t]{0.32\textwidth}
        \centering
        \includegraphics[width=\linewidth]{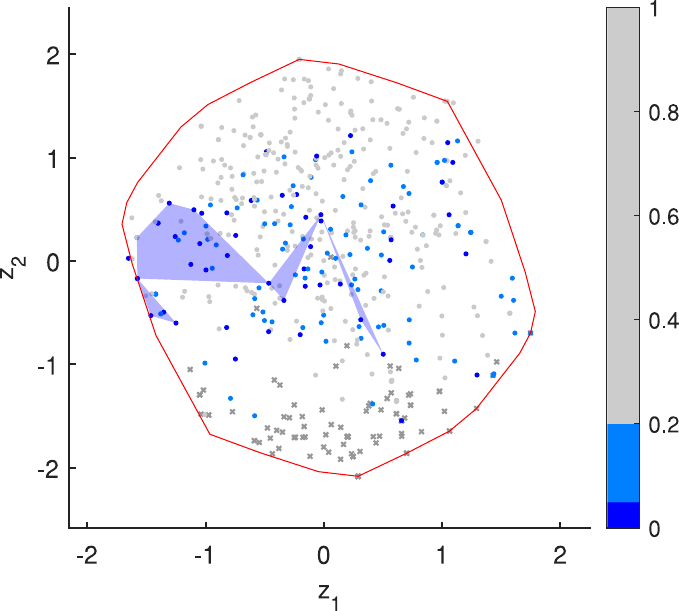}
	\caption{UoS}
	\label{fig:regret_UoS}
    \end{subfigure}
    \smallskip
    \begin{subfigure}[t]{0.32\textwidth}
        \centering
        \includegraphics[width=\linewidth]{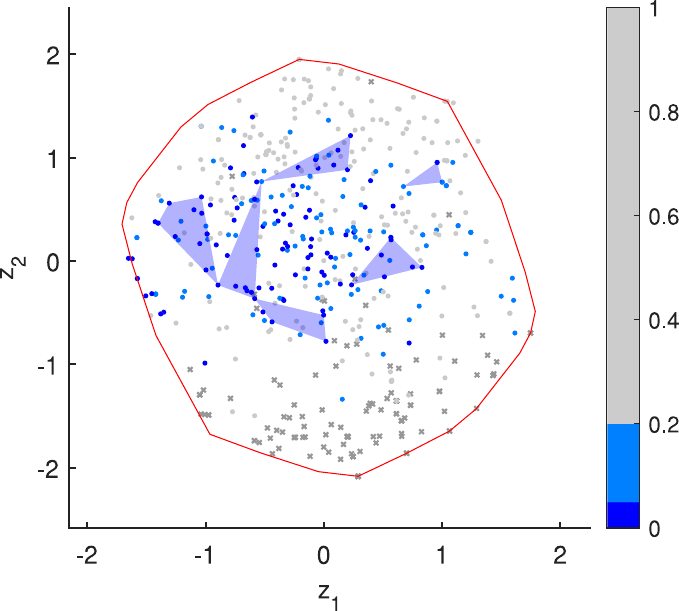}
	\caption{Goal}
	\label{fig:regret_Goal}
    \end{subfigure}
    \hfill
    \begin{subfigure}[t]{0.32\textwidth}
        \centering
        \includegraphics[width=\linewidth]{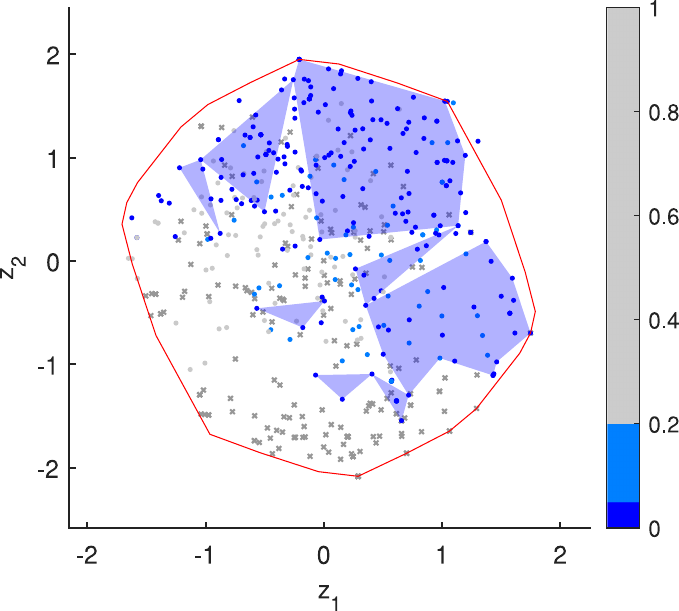}
	\caption{FBHS}
	\label{fig:regret_fbhs}
    \end{subfigure}
    \hfill
    \begin{subfigure}[t]{0.32\textwidth}
        \centering
        \includegraphics[width=\linewidth]{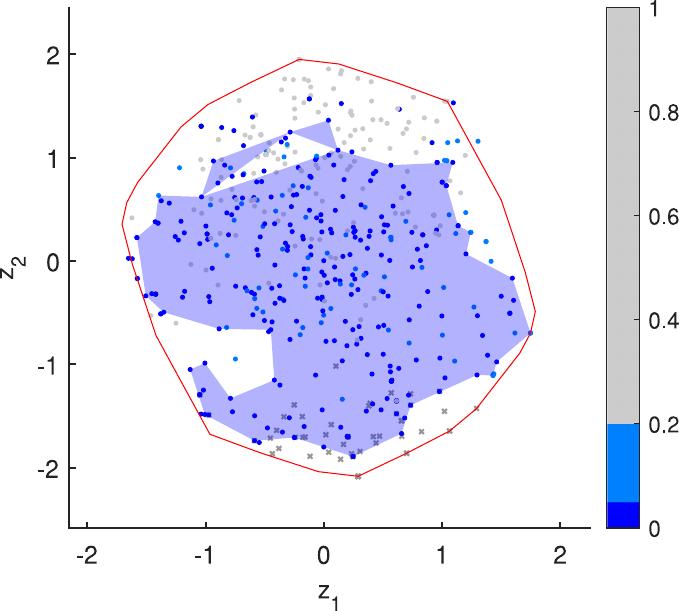}
	\caption{Udine}
	\label{fig:regret_udine}
    \end{subfigure}
    \smallskip
    \hfill
    \begin{subfigure}[t]{0.32\textwidth}
        \centering
        \includegraphics[width=\linewidth]{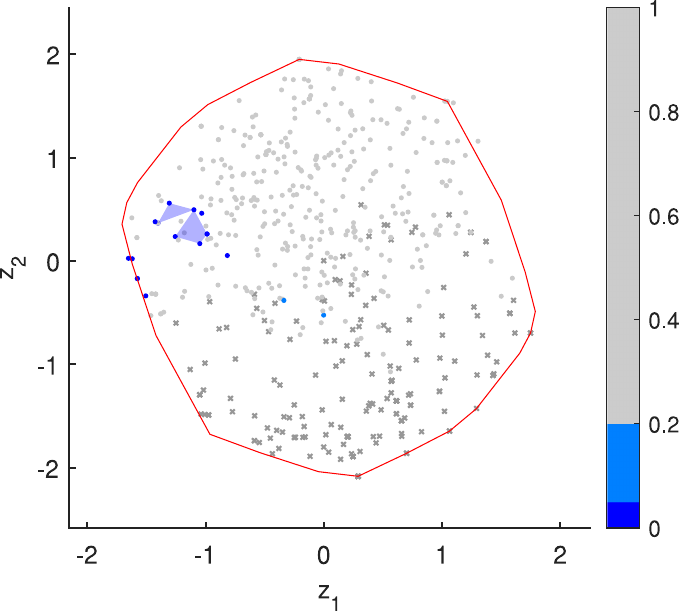}
        \caption{Reprobate}
	\label{fig:regret_uor}
    \end{subfigure}%
    \hfill
    \begin{subfigure}[t]{0.32\textwidth}
        \centering
        \includegraphics[width=\linewidth]{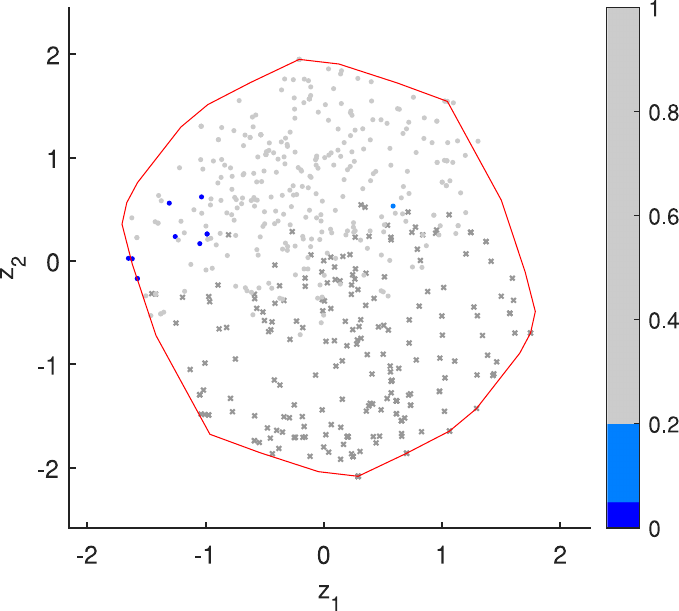}
	\caption{MODAL}
	\label{fig:regret_modal}
    \end{subfigure}
    \hfill
    \,
    \caption{Algorithm footprints. Problem instances for which no feasible solution was found are indicated by grey x-marks.}
    \label{fig:footprints}
\end{figure*}

In order to better appreciate the performance of each of the algorithms over the different problem instances, \Cref{fig:footprints} shows the relative gap of the algorithms in the 2D problem type space.
\new{
For each algorithm, the blue region in this figure corresponds to its so-called footprint, defined as the region in instance space for which the algorithm is expected to exhibit a strong performance (i.e., within 5\% of the best solution found). 
The ISA toolkit identifies this region by analysing the performance predictions on the training instances and then removing any regions with conflicting evidence.
As such, footprints provide an objective way to evaluate the relative strength of algorithms, considering the diversity of the problem instances as well as the strength of evidence (see \citet{Smith-Miles2023,Smith-Miles2012b}).
}

\new{The figure reveals that} Reprobate and MODAL behave quite similarly. Both struggle with finding best solutions. Indeed, even a clever combined approach of state-of-the-art solvers and use of a substantial computational effort results in no more than a handful of instances where MODAL tops the other algorithms. This suggests that complex sports scheduling problems are still beyond reach for IP solvers. MODAL also displays the lowest number of instances for which a feasible solution was found. Its footprint indicates that finding a feasible solution is particularly hard for problem instances in the bottom area of the problem type space. While Reprobate does not do better with respect to best solutions, it does manage to find a feasible solution for more problem instances, suggesting that constraint satisfaction techniques are more suited for this than IP models. Recall that Reprobate is based on pseudo-boolean (PB) optimization and that its main idea is to provide solutions with off-the-shelf generally applicable PB solvers. Nevertheless, Reprobate also fails in the region corresponding to problem instances that have a high value for the BR2 hard constraint. Considering an alternative PB constraint formulation for this constraint type may thus be a way to further improve this algorithm.

Although DITUoIArta and DES are rather different algorithms and DITUoIArta finds more feasible solutions than DES, the footprints of these algorithms are also quite similar.
Their footprints display excellent results in the middle left area, and occasional high-quality solutions along the 45$^{\circ}$ diagonal; beyond that region, both algorithms struggle. 
In fact, after the conclusion of the ITC2021 competition, this was illustrated by DES finding several new best solutions by employing considerably longer running times than used for the experiments of this paper.

The solvers UoS and Goal are alike in the sense that they are both matheuristics.
It is interesting to see that both of them perform particularly well along the middle-left to bottom-right diagonal of the problem type space.
None of the instance features seem to really dominate in this region, meaning that instances across this diagonal could be considered as `average instances'.
The two solvers still find a significant number of feasible solutions in other regions of the space, though, the quality of these solutions is somewhat less promising.

With regard to the FBHS solver, we observe that if a solution is found, it is very often close to the best known solution, especially when the number of breaks is to be minimized (BR2 soft constraint).
On the other hand, the solver clearly struggles to find feasible solutions in case the problem instances are phased or have a SE1 soft constraint, especially in combination with BR2 hard constraints.
As the FBHS solver constructs a timetable in two phases, it would be helpful to figure out whether the solver fails to find a feasible solution because the generated HAP set is infeasible, or rather because the matheuristic of the second phase fails to find a compatible opponent schedule even though the HAP set is in fact feasible.
In the former case, it would be interesting to see how the separation constraint and phased structure of the tournament can already (partially) be taken into account when constructing the HAP set.
In the latter case, replacing the matheuristic by other techniques like metaheuristics could help.

The footprint of Udine shows that the algorithm not only finds a feasible solution for the majority of the problem instances, but also that the solutions found are of high quality.
The footprint also demonstrates the effectiveness of the newly proposed phased neighbourhood by \citet{Rosati2022} as the solver performs very well in the part of the problem type space where phased problem instances are projected.
Only near the origin of the space, there is a gap where Goal and UoS sometimes seem to perform better; near the top of the space, the FBHS solver performs better.
Overall, the intuition is that the combination of neighbourhoods is the key element for the performance of the solver. To this regard, a promising direction of improvement could be the design of new neighbourhoods specifically aimed at reducing the number of breaks in the timetable, which might mitigate the feasibility issue near the bottom of the space. For example, the neighbourhood proposed by \citet{januario2016new} might be integrated in the solver. 

\Cref{tab:footprints} provides an overview of some metrics of the footprints: its area, which gives the area of good or even best performance relative to the area of the entire space, the density defined as the number of problem instances per unit area, and the purity which gives the percentage of instances enclosed by the footprint for which the algorithm provides a good or best solution (all three metrics should be as high as possible).
The algorithms with a significant footprint area are UoS, Goal, and especially FBHS and Udine. All four algorithms' footprints have a quite high density (Goal's footprint is even denser than the space on average), but the purity for Goal and UoS is only moderate meaning that there is some conflicting evidence of good and best performance in these areas (making performance predictions more challenging for those algorithms).
While Udine seems to succeed in finding a good solution for most of the problem instances, we observe that its footprint area for best algorithm performance is much smaller.
This suggest there is some room for algorithm selection techniques to recommend an appropriate algorithm in each part of the problem type space.

\begin{table}
    \centering
    \footnotesize
    \begin{tabular}{llllllllll}
        \toprule
	&& DES & DITUoIArta & UoS & Goal & FBHS & Udine & Reprobate & MODAL \\
        \midrule
        \multirow{3}{*}{Good} & Area & 0.018 & 0.019 & 0.101 & 0.112 & 0.579 & 0.956 & 0.01 & 0 \\
        & Density & 1.36 & 1.162 & 0.834 & 1.226 & 0.849 & 0.8 & 1.891 & 0 \\
	& Purity & 0.5 & 0.778 & 0.6 & 0.632 & 0.737 & 0.73 & 0.75 & 0 \\[7pt]
        \multirow{3}{*}{Best} & Area & 0 & 0 & 0.03 & 0.019 & 0.247 & 0.399 & 0 & 0 \\
        & Density & 0 & 0 & 0.883 & 1.657 & 0.883 & 0.758 & 0 & 0 \\
        & Purity & 0 & 0 & 0.727 & 0.615 & 0.769 & 0.786 & 0 & 0 \\
        \bottomrule
    \end{tabular}
    \caption{Algorithm footprint metrics based on the area of good and best performance of an algorithm \new{in the problem type space} (training instances only)}
    \label{tab:footprints}
\end{table}

\subsubsection{Computation time}

\begin{figure}
    \begin{subfigure}[t]{0.49\textwidth}
	\includegraphics[width=0.95\linewidth]{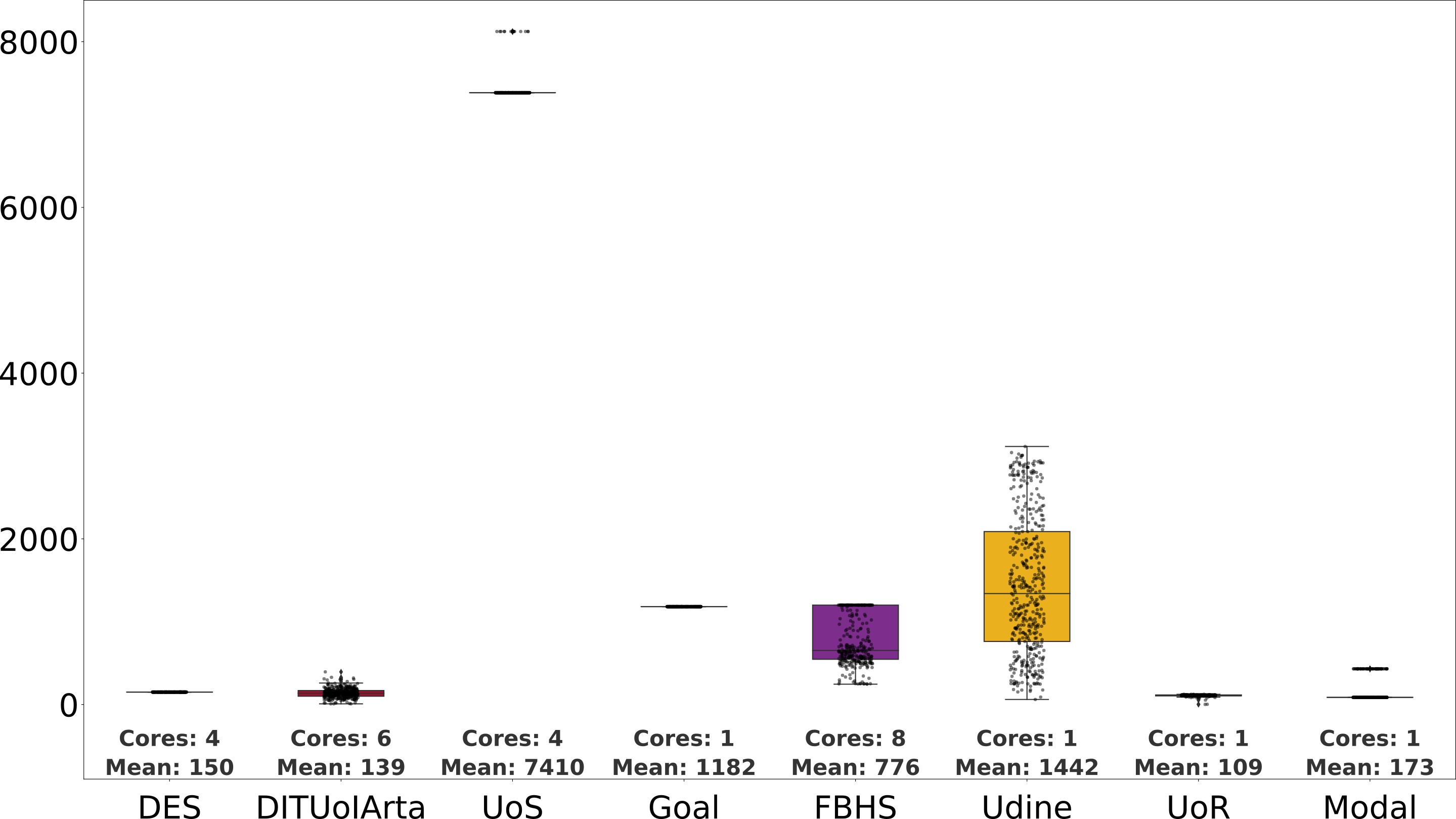}
	\caption{Wall time (minutes)}
    \end{subfigure}
    \hfill
    \begin{subfigure}[t]{0.49\textwidth}
	\includegraphics[width=0.95\linewidth]{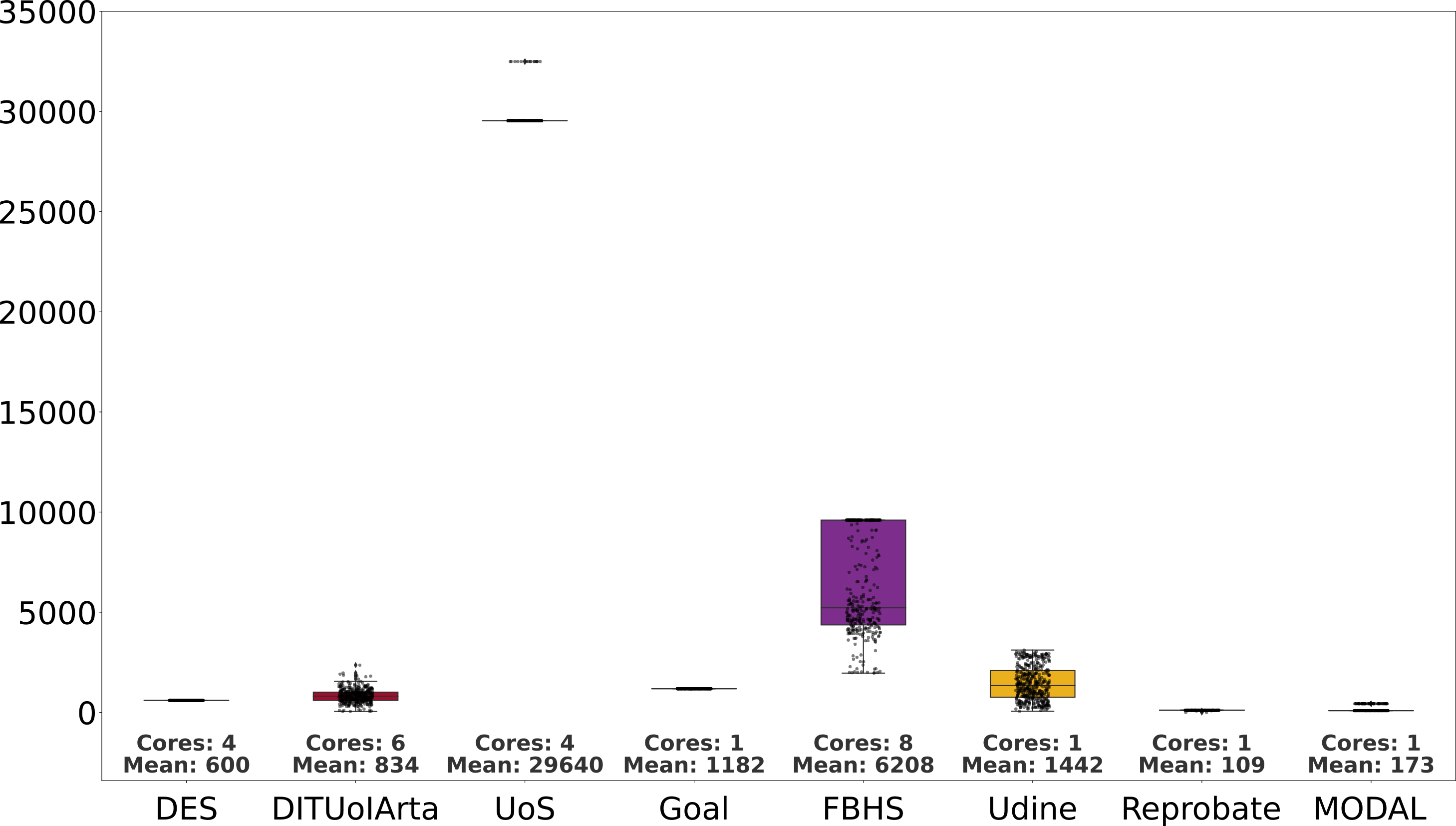}
	\caption{CPU time (minutes)}
    \end{subfigure}
    \caption{Distribution of the runtimes for the additional problem instances per algorithm. For each algorithm, only instances for which it found a solution are considered. Runtimes are normalized with regard to processor clock speed by multiplying with the clock speed ratio from \Cref{tab:algImpl}.}
	\label{fig:runtimeBox}
\end{figure}

\Cref{fig:runtimeBox} provides an overview of the normalized  wall times (i.e., the total execution time of the code) and CPU times (i.e., wall time multiplied with the number of cores used) utilized by the various algorithms. The figure shows that wall times can be categorized into three groups: short (up to several hours: DES, DITUoIArta, Reprobate, and MODAL), medium (about a day: Goal, FBHS, Udine), and long (several days: UoS). Some of these differences are substantial, however, as we pointed out before, the quality of the sports timetable is typically considered much more important than the speed of computation (see \citet{VanBulck2022b}).

\begin{figure*}
    \centering
    \begin{subfigure}[t]{0.32\textwidth}
        \centering
        \includegraphics[width=\linewidth]{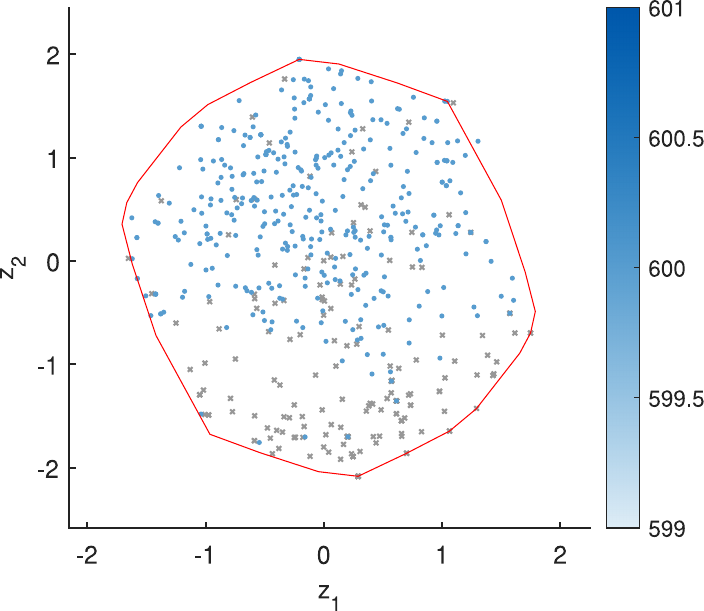}
	\caption{DES}
	\label{fig:runtime_des}
    \end{subfigure}
    \hfill
    \begin{subfigure}[t]{0.32\textwidth}
        \centering
        \includegraphics[width=\linewidth]{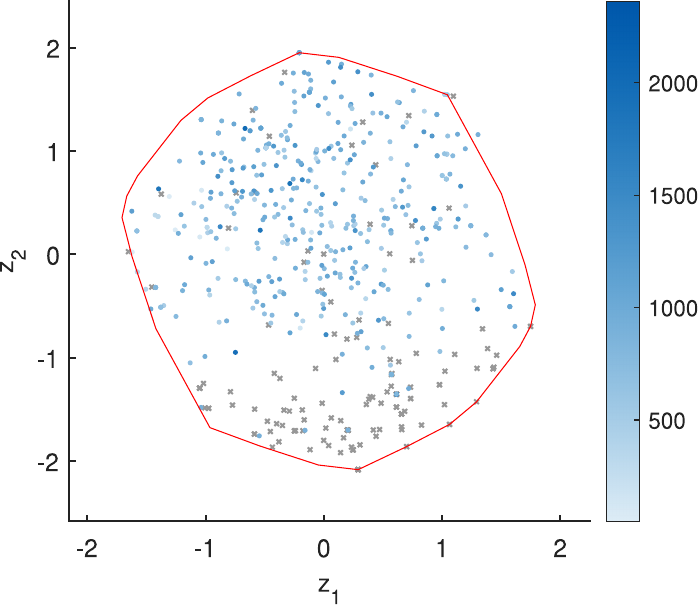}
	\caption{DITUoIArta}
	\label{fig:runtime_arta}
    \end{subfigure}
    \hfill
    \begin{subfigure}[t]{0.32\textwidth}
        \centering
        \includegraphics[width=\linewidth]{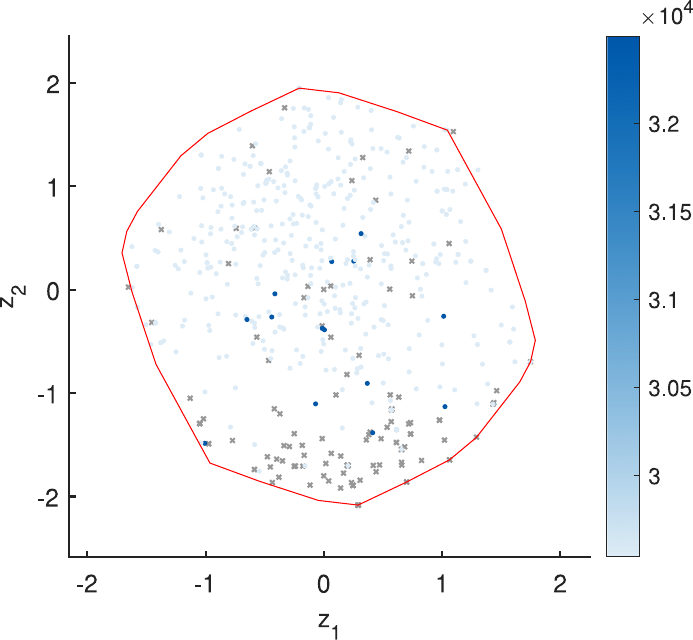}
	\caption{UoS}
	\label{fig:runtime_UoS}
    \end{subfigure}
    \smallskip
    \begin{subfigure}[t]{0.32\textwidth}
        \centering
        \includegraphics[width=\linewidth]{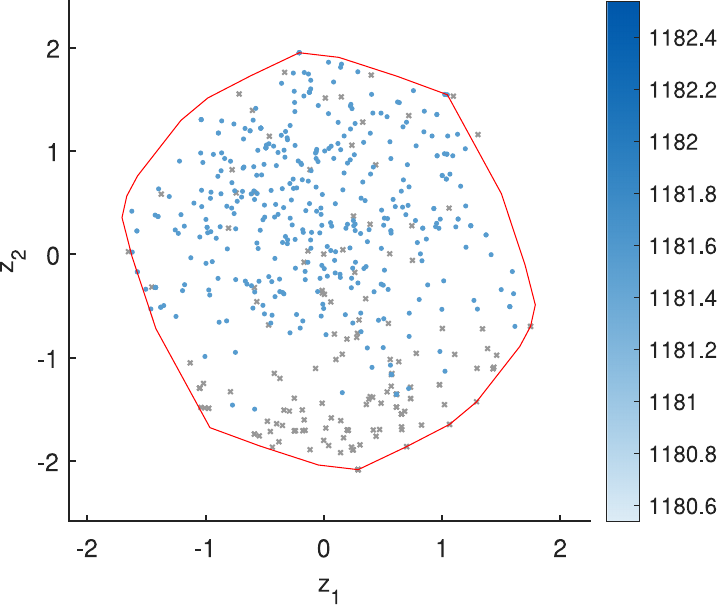}
	\caption{Goal}
	\label{fig:runtime_Goal}
    \end{subfigure}
    \hfill
    \begin{subfigure}[t]{0.32\textwidth}
        \centering
        \includegraphics[width=\linewidth]{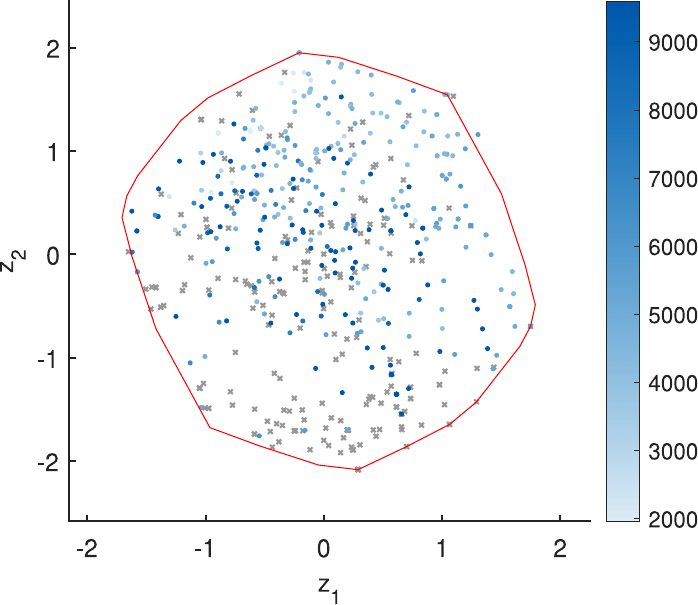}
	\caption{FBHS}
	\label{fig:runtime_fbhs}
    \end{subfigure}
    \hfill
    \begin{subfigure}[t]{0.32\textwidth}
        \centering
        \includegraphics[width=\linewidth]{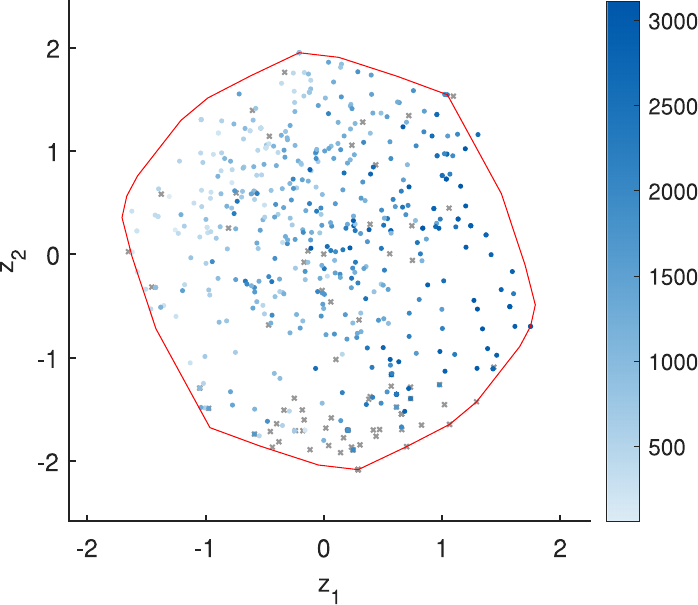}
	\caption{Udine}
	\label{fig:runtime_udine}
    \end{subfigure}
    \smallskip
    \hfill
    \begin{subfigure}[t]{0.32\textwidth}
        \centering
        \includegraphics[width=\linewidth]{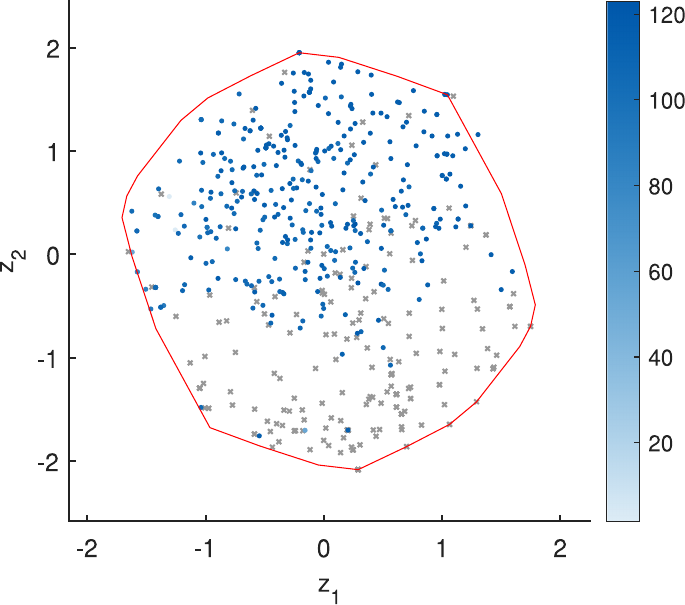}
        \caption{Reprobate}
	\label{fig:runtime_uor}
    \end{subfigure}%
    \hfill
    \begin{subfigure}[t]{0.32\textwidth}
        \centering
        \includegraphics[width=\linewidth]{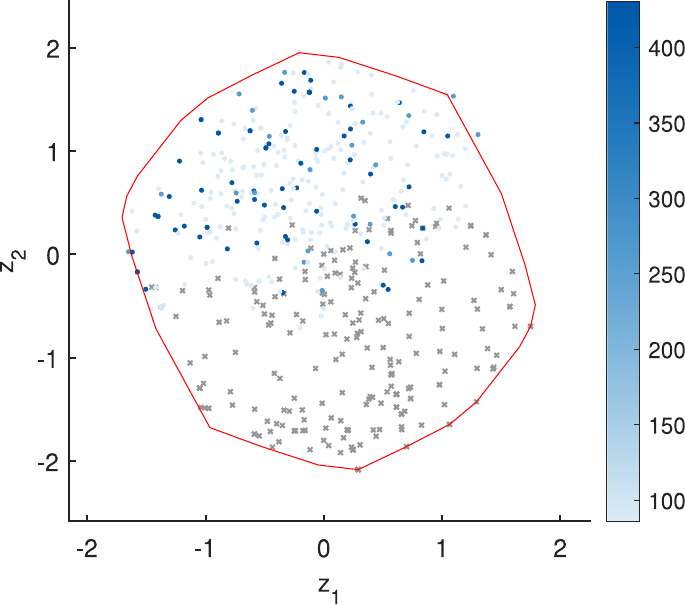}
	\caption{MODAL}
	\label{fig:runtime_modal}
    \end{subfigure}
    \hfill
    \,
    \caption{Distribution of wall times (in minutes) over the problem type space. Problem instances for which no feasible solution was found are indicated by a grey x-mark.}
    \label{fig:runTimes}
\end{figure*}

A more detailed view of the distribution of wall times over the problem instances is given in \Cref{fig:runTimes}.
While it can be seen that DES, UoS (except for few problem instances), Goal, Reprobate, and to some extent DITUoIArta feature more or less uniform runtimes for all considered problem instances, we observe some interesting patterns for FBHS and Udine.
First, with regard to FBHS, we see that runtimes are longer for problem instances that are phased or have a SE1 constraint.
Perhaps, this can be best explained by the reduced flexibility to construct a compatible opponent schedule for such problem instances.
Also interesting to note is that runtimes are shorter in these regions where the FBHS solver excels.
Second, with regard to the solver by Udine, we observe that runtimes tend to increase from left to right in the problem type space.
We note that this trend corresponds to an increase of FA2 soft, CA2 hard and CA3 hard constraints. 
As the solver by Udine is evaluation based, it would be interesting to see whether the increase in runtime can be linked to, e.g., expensive computations of the change in costs for these constraints after applying a move.

\subsection{Instance versus problem type space}

In order to investigate the impact of working with problem type rather than instance-specific features, we also used the ISA toolkit to construct a problem instance space based on \new{all problem type features (\Cref{tab:problemTypeFeatures}) plus all instance-specific features (\Cref{tab:instanceFeatures})}. 
The projection matrix, again based on the training instances only, for this instance space is given by the following equation which also shows the final selection of features.

\begin{singlespace}
{
	\small
\begin{equation}
	\bm{(z_1,z_2)} = 
	\begin{bmatrix} 
		-0.2398		& -0.1241 	\\[2pt]
		\phantom{-}0.1657		& -0.1184 	\\[2pt]
		\phantom{-}0.2899		& -0.3249 	\\[2pt]
		\phantom{-}0.4703		& -0.0171 	\\[2pt]
		\phantom{-}0.2474		& -0.1614 	\\[2pt]
		-0.358		& -0.2934 	\\[2pt]
		-0.0997		& \phantom{-}0.4328 	\\[2pt]
		\phantom{-}0.5187		& -0.1947 	\\[2pt]
		-0.2241		& -0.1807 	\\[2pt]
		\phantom{-}0.7985		& \phantom{-}0.0181 	\\[2pt]
		-0.1027		& -0.0802 	\\[2pt]
		\phantom{-}0.471		& -0.2456 	\\
	\end{bmatrix}^{\intercal}
	\begin{bmatrix*}[l]
		f_{|T|}\\
		f^S_{\text{CA1}}\\[2pt]
		f^S_{\text{CA2}}\\[2pt]
		f^H_{\text{CA3}}\\[2pt]
		f^H_{\text{CA4}}\\[2pt]
		f^S_{\text{CA4}}\\[2pt]
		f^S_{\text{BR2}}\\[2pt]
		\phi^{IP}_{\text{obj mean}}\\[2pt]
		\phi^{IP}_{\text{cons degree max}}\\[2pt]
		\phi^{IP}_{\text{LP objective}}\\[2pt]
		\phi^{SA}_{\text{CA3}}\\[2pt]
		\phi^{SA}_{\text{soft cost stage 1 2}}\\
	\end{bmatrix*}
\end{equation}
}
\end{singlespace}

\begin{figure}[tp]
	\hfill
    	\begin{subfigure}[t]{0.4\linewidth}
        	\includegraphics[width=\linewidth]{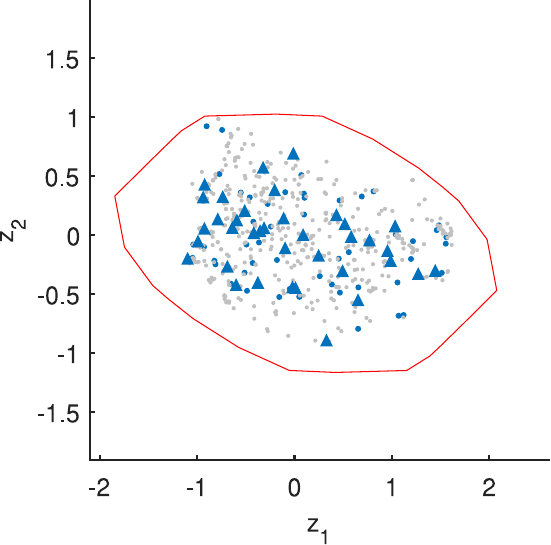}
            \caption{\new{Problem instances}}
		\label{fig:instanceSpaceInstances}
	\end{subfigure}
	\hfill
    	\begin{subfigure}[t]{0.45\linewidth}
        	\includegraphics[width=\linewidth]{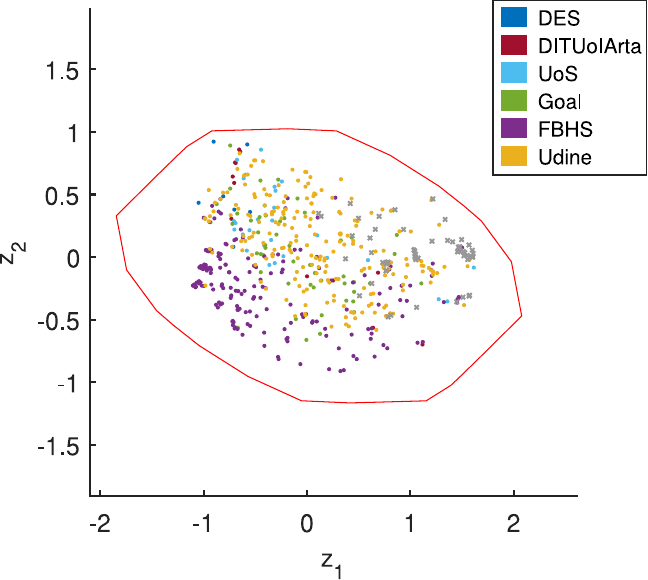}
         \caption{\new{Best algorithm}}
		\label{fig:instanceSpaceAlgorithms}
	\end{subfigure}
	\hfill
	\,
	\caption{Projections of the problem instances \textit{(a)} and best performing algorithm \textit{(b)} in the problem instance space constructed with instance-specific features. 		Grey and blue circles and blue triangles represent the set of training, validation, and ITC2021 problem instances, respectively. }
	\label{fig:lowLevelSpace}
\end{figure}

\Cref{fig:lowLevelSpace} illustrates the distribution of training and test instances along with the best-performing algorithm for each of these instances.
Despite the inability of our instance generator to directly influence instance-specific features, \Cref{fig:instanceSpaceInstances} suggests that the generated problem instances still encompass a significant portion of the overall instance space. 
This not only emphasizes the diversity of our instance benchmark but also suggests that establishing a diverse benchmark at the level of the problem types is a valuable step towards diversity in terms of instance-specific features.
\new{Comparing \Cref{fig:numberBest,fig:instanceSpaceAlgorithms}, we note a visually improved separation between the regions of best performance for the FBHS solver and Udine. 
This translates into a higher purity of the footprints, as noticed by comparing \Cref{tab:footprints,tab:footprints_problemtype}.} 
\new{Perhaps, the separation could be further enhanced by devising more sophisticated instance-specific features tailored towards sports timetabling.}
However, such features are \new{typically} less intuitive \new{(see also \citet{DeCoster2021})} and considerably more challenging to predict without knowledge of the specific details of the problem instances to be addressed in the future.
\new{Moreover, even though these features may help to better predict the performance of algorithms, they do not necessarily help us to explain why this is the case (see also \citet{Smith-Miles2012}).}
\new{The cost of working with features at the level of problem types, in terms of the quality of algorithm performance predictions, is discussed in the next section.}


\begin{table}
    \centering
    \footnotesize
    \begin{tabular}{llllllllll}
        \toprule
	&& DES & DITUoIArta & UoS & Goal & FBHS & Udine & Reprobate & MODAL \\
        \midrule
\multirow{3}{*}{Good} & Area &0.007	&0.015	&0.036	&0.105	&0.332	&0.64	&0.009	&0\\
&Density    &2.266	&1.833	&1.262	&1.316	&1.184	&1.021	&1.413	&0\\
&Purity     &0.667	&0.818	&0.778	&0.709	&0.828	&0.835	&0.8	&0\\[7pt]

\multirow{3}{*}{Best} &Area &0	&0	&0.003	&0.027	&0.288	&0.363	&0	&0\\
& Density                   &0	&0	&3.414	&1.135	&0.983	&1.117	&0	&0\\
&Purity                     &0	&0	&0.75	&0.667	&0.876	&0.759	&0	&0\\
        \bottomrule
    \end{tabular}
    \caption{\new{Algorithm footprint metrics based on the area of good and best performance of an algorithm in the instance space (training instances only)}}
    \label{tab:footprints_problemtype}
\end{table}

\subsection{Automated algorithm recommendations}

The previous two subsections revealed that there is no single solver that performs best on all problem instances, and that depending on the location in problem type or instance space, some algorithms may work better than others.
This section therefore investigates whether we can develop algorithm selection techniques to predict which algorithm is most suitable for each of the problem instances.
To come up with these recommendations, we make use of the build-in-functionality of the ISA toolkit and of the algorithm selection software AutoFolio (see \citet{Lindauer2015}).

\begin{figure}
    \hfill
    \begin{subfigure}[t]{0.475\textwidth}
        \centering
        \includegraphics[width=\linewidth]{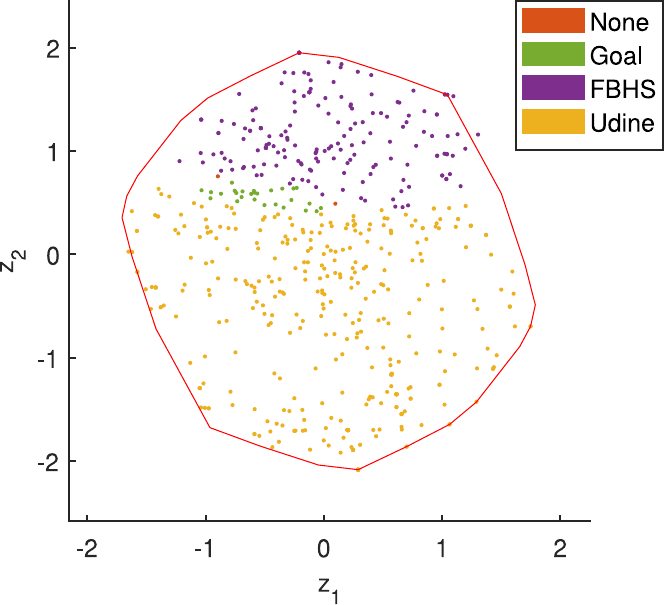}
	\caption{Recommendations}
    \end{subfigure}
    \hfill
    \begin{subfigure}[t]{0.475\textwidth}
        \centering
        \includegraphics[width=\linewidth]{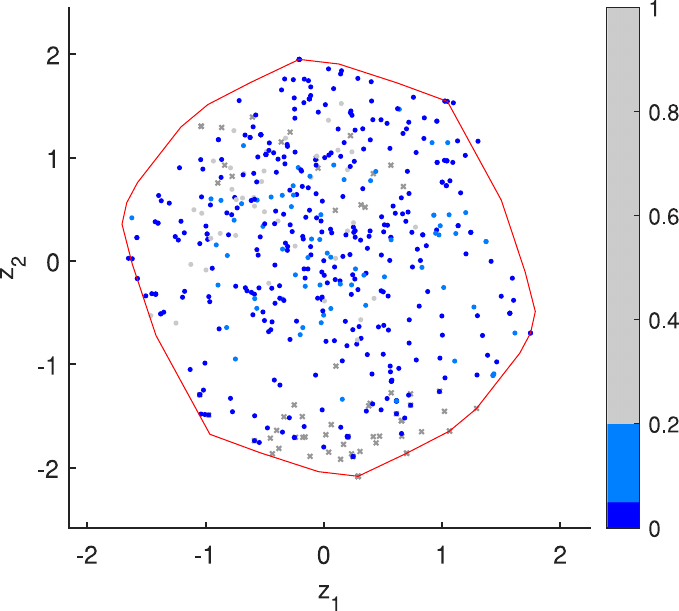}
	\caption{Performance}
    \end{subfigure}
    \hfill
    \,
    \smallskip
    \hfill
    \begin{subfigure}[t]{0.475\textwidth}
        \centering
        \includegraphics[width=\linewidth]{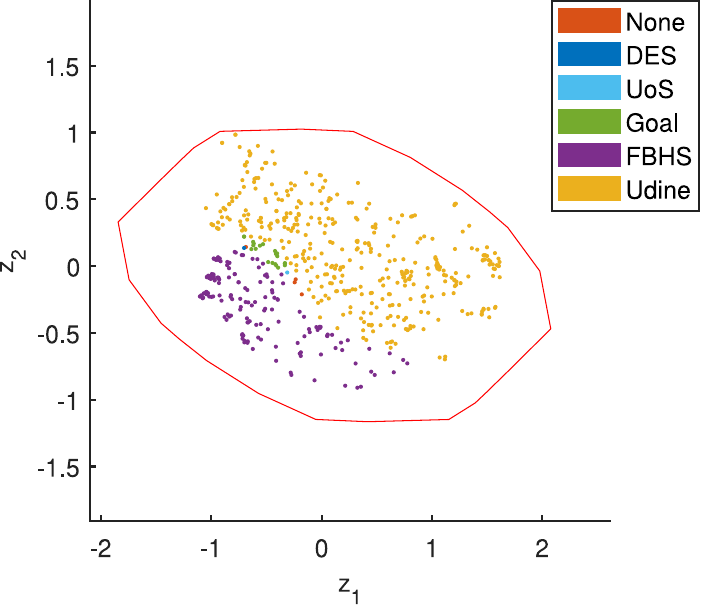}
	\caption{Recommendations}
    \end{subfigure}
    \hfill
    \begin{subfigure}[t]{0.475\textwidth}
        \centering
        \includegraphics[width=\linewidth]{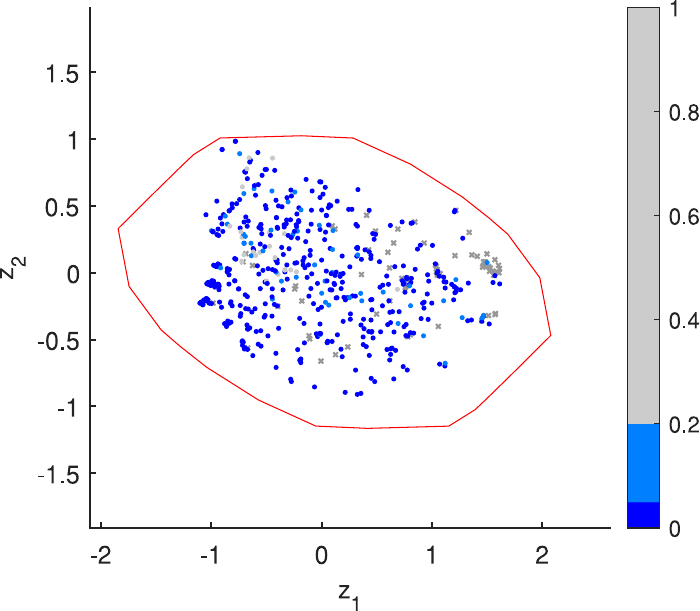}
	\caption{Performance}
    \end{subfigure}
    \hfill
    \,
    \caption{
	    Algorithm recommendations by the ISA toolkit and performance thereof using as features the two-dimensional coordinates in the problem type (a and b) and instance space (c and d).
    }
	\label{fig:bestAlgo}
\end{figure}

Based on the projection coordinates in either the 2D problem type or the 2D instance space, the ISA toolkit utilizes Support Vector Machines (SVM) to predict whether each algorithm achieves good performance.
A visualization of these recommendations and the associated performance of the recommended algorithms is given in \Cref{fig:bestAlgo}.
The figure reveals that the SVM model rarely makes mistakes near the top and bottom of the spaces; \Cref{fig:numberFeasBest,fig:lowLevelSpace} show that Udine and FBHS clearly dominate in these regions.
On the other hand, more mistakes are made near the middle of the instance space, where multiple algorithms produce best solutions.
Though, only for few instances, these mistakes result in a substantial gap.

In addition to the ISA toolkit, we also make use of AutoFolio which is a highly parametrizable algorithm selection software tool. Thanks to its build-in parameter tuning, AutoFolio can be applied out-of-the-box to novel algorithm selection applications (see \citet{Lindauer2015}).
Among others, parameter choices include options like XGBoost or Random Forests as the machine learning model, and choosing between pairwise classification (like in the ISA toolkit) or regression.
In the latter, a model is trained for each algorithm, and the one with the lowest predicted relative gap is chosen. 
Rather than basing predictions on the coordinates in the 2D spaces, we use AutoFolio to make predictions based on the full set of problem type features, both with and without augmentation by the set of instance-specific features from \Cref{tab:instanceFeatures}.

\Cref{tab:algoRes} provides the performance of the algorithm recommendation systems on the test set in terms of the percentage of feasible, best, and good solutions as well as the average relative gap. 
In addition, the table also provides these metrics for the single best algorithm, which is the algorithm with the highest overall performance, and an oracle always predicting the best performing algorithm. 
In our case, we select Udine as the single best solver as it found most of the feasible and good solutions and has the lowest average relative gap. 
However, it is worth noting that the percentage of best solutions is slightly higher for the FBHS solver compared to the Udine solver (43\% vs. 39\%). 
Examining the recommendation systems' performance reveals a comparable number of feasible solutions, but a significant enhancement in best and good solutions compared to the single best solver. 
Notably, the recommendation systems exhibit improvements of up to 30\% in the percentage of best and good solutions, along with an impressive 10\% enhancement in the relative gap. 
This confirms that combining the strengths of diverse algorithms through recommendation systems results in an overall performance boost for the given problem instances.
A comparison of the performance of the recommendation systems in the problem type versus instance space using 2D coordinates indicates an increase of nearly 10\% in best and good solutions while no significant changes in the relative gap are found.
These findings are encouraging, considering that the features in the problem type space are more intuitive and easier to predict. Moreover, when utilizing the complete set of feature values (as in AutoFolio) instead of just projection coordinates (as in the ISA toolkit), minimal disparities in performance results are observed, with AutoFolio performing better in the problem type than in the instance space, a trend attributed to randomness in the data.

\begin{table}
	\footnotesize
 \centering
\begin{tabular}{l c ll c ll c l c l}
\toprule
			&&\multicolumn{2}{c}{Problem type space} && \multicolumn{2}{c}{Problem instance space}\\
			\cmidrule{3-4} \cmidrule{6-7}
			&& ISA & Autofolio && ISA & Autofolio && Single Best (Udine) && Oracle\\
\midrule                                                                                                                     
		Percentage feas.\ 		&& 97.6\% & 100\% 	&& 98.8\% & 100\% 	&& 100\% 	&& 100\%\\
		Percentage best 	&& 54\% & 70\% 		&& 64\% & 64\% 		&& 39\% 	&& 100\%\\
		Percentage good 	&& 71\% & 83\% 		&& 79\% & 79\% 		&& 53\% 	&& 100\%\\
		Relative gap  		&& 4.93\% & 2.93\% 	&& 4.11\% & 3.49\%  	&& 12.8\% 	&& 0\%\\
		\bottomrule
\end{tabular}
	\caption{Performance metrics for the various algorithm recommendation systems.}
  \label{tab:algoRes}
\end{table}

\subsection{Analysis of algorithm complementarity}
\label{subsec:compl}

The previous sections have shown that sports timetabling algorithms have different strengths and weaknesses, allowing machine learning techniques to construct a portfolio of algorithms that together performs better than its individual components.
An interesting question is to what extent each of the algorithms contributes to the strength of the portfolio.
One possibility to measure an algorithm's contribution is its standalone performance in terms of the relative gap. 
However, this metric completely overlooks the complementarity of solvers. 
For instance, a perfect clone of an algorithm would receive exactly the same score, although clearly not adding any value to the portfolio.
A second possibility is to calculate the marginal performance of an algorithm, which is the difference in performance between  the portfolio after removing the algorithm and the portfolio of all algorithms. 
In other words, the marginal performance indicates the improvement in the relative gap when a new algorithm becomes available to an error-free algorithm recommendation system (see, e.g., \citet{Wagner2018}).
Table \ref{tab:shapley} shows that Udine and FBHS receive the highest marginal score, which is possibly due to the fact that Udine provides several solutions to otherwise unsolved problem instances and that FBHS is able to improve upon many solutions.
All other solvers receive very low scores, likely due to their strongly correlated performance. 
Nonetheless, having at least one of these algorithms in the portfolio is desirable and should therefore be rewarded.
To that purpose, \Cref{tab:shapley} also provides Shapley scores which are equal to the marginal performance of an algorithm with respect to all possible algorithm (sub)portfolios (see \citet{Frechette2016}). 
As \Cref{tab:shapley} shows, the strength of the algorithm portfolio not only comes from Udine and FBHS, but also from solvers like DITUoIArta, UoS, and Goal, which on average improve the relative gap of a portfolio by about 13 percent.

\begin{table}
	\footnotesize
 \centering
\begin{tabular}{l l l l l l l l l}
\toprule
              		& DES 	& DITUoIArta 	& UoS 	& Goal 	& FBHS & Udine & Reprobate  & MODAL	   	\\
\midrule                                                                                                                     
	Standalone  	& 58.81\% & 49.40\% & 35.94\% & 35.81\% & 38.15\% & 12.83\% & 73.34\% & 74.06\% \\
	Marginal    	& 0\% & 0.07\% & 0.89\% & 0.66\% & 9.7\% & 11.56\% & 0\% & 0\%\\
	Shapley     	& 6.28\% & 8.59\% & 13.63\% & 13.58\% & 21.16\% & 29.38\% & 3.78\% & 3.6\%\\
\bottomrule
\end{tabular}
	\caption{Standalone, marginal, and Shapley scores in terms of the relative gap. If no solution was found, the relative gap is set equal to 1.}
  \label{tab:shapley}
\end{table}

\section{Conclusion}
\label{sec:conclusion}

\new{The performance of (sports) timetabling algorithms is still too often based on a comparison against manually generated solutions. The current paper provides a viable alternative for sports timetabling, using techniques from instance space analysis to properly assess the performance of sports timetabling algorithms on a diverse set of problem instances. Our results show 
\begin{paraenum}
    \item that no single algorithm performs best on all problem instances, and 
    \item that based on some high-level information of the number of constraints of each type, high-quality predictions can be made about which algorithm is expected to perform best. 
\end{paraenum}
    This offers researchers insights on the strengths and weaknesses of their algorithms, and provides practitioners with an idea of which algorithm to use to schedule their sports tournament. We anticipate that this study triggers further improvements on the instance benchmark, which can be tracked on itc2021.ugent.be. Indeed, this benchmark remains challenging, knowing that so far only two instances have been solved to optimality.
}


As an area for future work, we like to point out that algorithms in a portfolio can add value beyond the situation where we might select them as the best algorithm to run. At the same time, this serves as a caveat to our analysis of algorithm complementarity. Indeed, if algorithms are usable in conjunction (i.e.\ the output of one feeding into another, forming a composite algorithm), an individual algorithm could substantially improve the overall performance, even if weak as a standalone. This was demonstrated by the DES algorithm at the conclusion of ITC2021. Warm-starting on the best-known solution for each instance (effectively generated by a portfolio of all other algorithms), it was able to improve a third of these solutions relatively quickly (see \citet{Phillips2021}). Likewise, given the complementary performances of the Udine and the FBHS solver, it seems a promising line of research to see whether \new{a} hybrid approach, warm starting the Udine solver with the best solution found by the FBHS solver (perhaps ran with a shorter computation time, and possibly violating some of the hard constraints) will lead to further performance improvements for this challenging problem.

\new{
Finally, we point out that there are other timetabling and scheduling applications that suffer from the problem that algorithms described in the literature have mostly been compared with manually created solutions for a limited set of (often undisclosed) problem instances, obscuring the assessment of algorithm performance. We hope that our approach, including the focus on higher level problem-type features, may inspire researchers in these domains.
}

\vspace{-20pt}
\begin{singlespace}
\section*{In memoriam}
We dedicate this article to Jan-Patrick Clarner, who passed away unexpectedly during this research. He wrote his last email about where we could continue his work.
\end{singlespace}

\vspace{-20pt}
\begin{singlespace}
\section*{CRediT authorship contribution statement}
\new{
\textbf{David Van Bulck:} Conceptualization, Data curation, Formal analysis, Investigation, Validation, Visualization, Writing – original draft, 
Writing – review \& editing.
\textbf{Dries Goossens:} Conceptualization, Project administration, Writing – original draft, Writing – review \& editing.
\textbf{Jan-Patrick Clarner, Angelos Dimitsas, George H.\ G.\ Fonseca, Carlos Lamas-Fernandez, Martin Mariusz Lester, Jaap Pedersen, Antony E. Phillips, Roberto Maria Rosati:} Methodology, Software, Writing – original draft, Writing – review \& editing.
}
\end{singlespace}

\vspace{-20pt}
\begin{singlespace}
\section*{Acknowledgement}
David Van Bulck is a postdoctoral research fellow funded by the Research Foundation – Flanders (FWO) [1258021N]. Roberto Maria Rosati acknowledges the CINECA award under the ISCRA initiative, for the availability of high-performance computing resources and support.
\end{singlespace}
\vspace{-20pt}



\begin{singlespace}


\end{singlespace}

\end{document}